\documentclass[10pt,twocolumn,letterpaper]{article}

\usepackage{iccv}
\usepackage[utf8]{inputenc}
\usepackage{times}
\usepackage{epsfig}
\usepackage{graphicx}
\usepackage{amsmath,amsfonts,amssymb,amsthm}
\usepackage{color}
\usepackage{cite}
\usepackage{array}
\usepackage{xspace}
\usepackage{caption}
\usepackage{subcaption}
\usepackage{enumitem}
\usepackage{booktabs}
\usepackage{multirow}
\usepackage[normalem]{ulem}
\usepackage[toc,page]{appendix}
\usepackage[accsupp]{axessibility}

\usepackage[pagebackref=true,breaklinks=true,colorlinks,bookmarks=false]{hyperref}

\newcommand{\figref}[1]{Fig.~\ref{#1}}
\newcommand{\secref}[1]{Section~\ref{#1}}

\newcommand{\tabref}[1]{Table~\ref{#1}}

\makeatletter
\DeclareRobustCommand\onedot{\futurelet\@let@token\@onedot}
\def\@onedot{\ifx\@let@token.\else.\null\fi\xspace}
\def\eg{e.g\onedot} 
\def\ie{i.e\onedot}

\makeatother

\newcommand{\boldparagraph}[1]{\vspace{0.0cm}\noindent{\bf #1:} }

\definecolor{darkgreen}{rgb}{0,0.7,0}
\definecolor{darkblue}{RGB}{31,119,180}
\definecolor{darkred}{RGB}{214,39,40}
\definecolor{mediumgray}{rgb}{0.5,0.5,0.5}
\definecolor{mediumteal}{rgb}{0,0.5,0.5}

\definecolor{ellisred}{rgb}{0.87,0.44,0.38} %
\definecolor{ellisgreen}{rgb}{0.69,0.90,0.52} %
\definecolor{elliscyan}{rgb}{0.29,0.77,0.74} %
\definecolor{ellisorange}{rgb}{0.89,0.55,0.28} %
\definecolor{ellisblue}{rgb}{0.41,0.61,0.86} %

\usepackage{appendix}

\newcommand{\pmsd}[1]{{\color{mediumgray}{\scriptsize $\pm$ #1}}}

\iccvfinalcopy %

\def\iccvPaperID{1323} %

\ificcvfinal\fi

\begin{document}

\title{Hidden Biases of End-to-End Driving Models}

\author{
Bernhard Jaeger \quad Kashyap Chitta \quad Andreas Geiger\\
University of Tübingen \quad Tübingen AI Center\\
{\tt\small \{bernhard.jaeger, kashyap.chitta, a.geiger\}@uni-tuebingen.de}
}

\maketitle
\ificcvfinal\thispagestyle{empty}\fi

\begin{abstract}
   End-to-end driving systems have recently made rapid progress, in particular on CARLA. Independent of their major contribution, they introduce changes to minor system components. Consequently, the source of improvements is unclear. We identify two biases that recur in nearly all state-of-the-art methods and are critical for the observed progress on CARLA: (1) lateral recovery via a strong inductive bias towards target point following, and (2) longitudinal averaging of multimodal waypoint predictions for slowing down. We investigate the drawbacks of these biases and identify principled alternatives. By incorporating our insights, we develop TF++, a simple end-to-end method that ranks first on the Longest6 and LAV benchmarks, gaining 11 driving score over the best prior work on Longest6.
\end{abstract}
\vspace{-0.8cm}

\section{Introduction}

End-to-end driving approaches have rapidly improved in performance on the CARLA leaderboard~\cite{Leaderboard2020}, the de-facto standard for fair online evaluation. Driving scores have increased from under 20~\cite{Chen2019CORL, Prakash2021CVPR} to over 70~\cite{Shao2022CORL, Wu2022NeurIPS} in just two years. However, why recent systems work so well is not fully understood, as both methods and training sets differ largely between submissions. Rigorous ablations are expensive due to the large design space of driving systems and the need to simulate large amounts of driving for evaluation.

\begin{figure}[t]
\centering
    \begin{subfigure}[b]{0.475\linewidth}
        \centering
        \includegraphics[width=\textwidth]{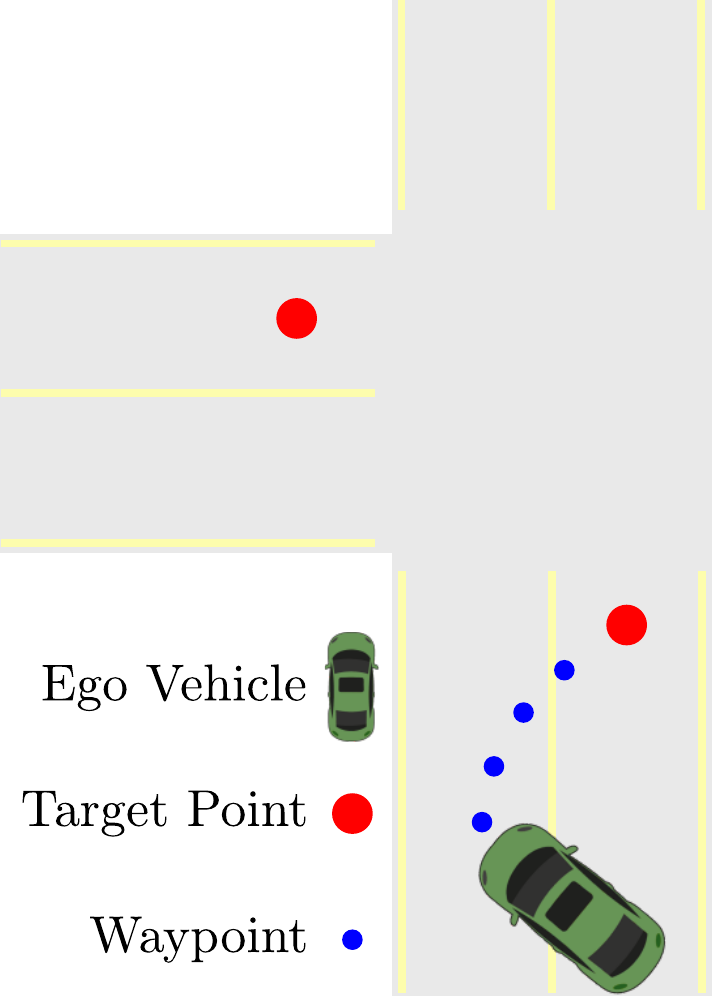}
        \caption{Target point shortcut}
        \label{fig:teaser_1}
    \end{subfigure}
    \hfill
    \begin{subfigure}[b]{0.475\linewidth}
        \centering
        \includegraphics[width=\textwidth]{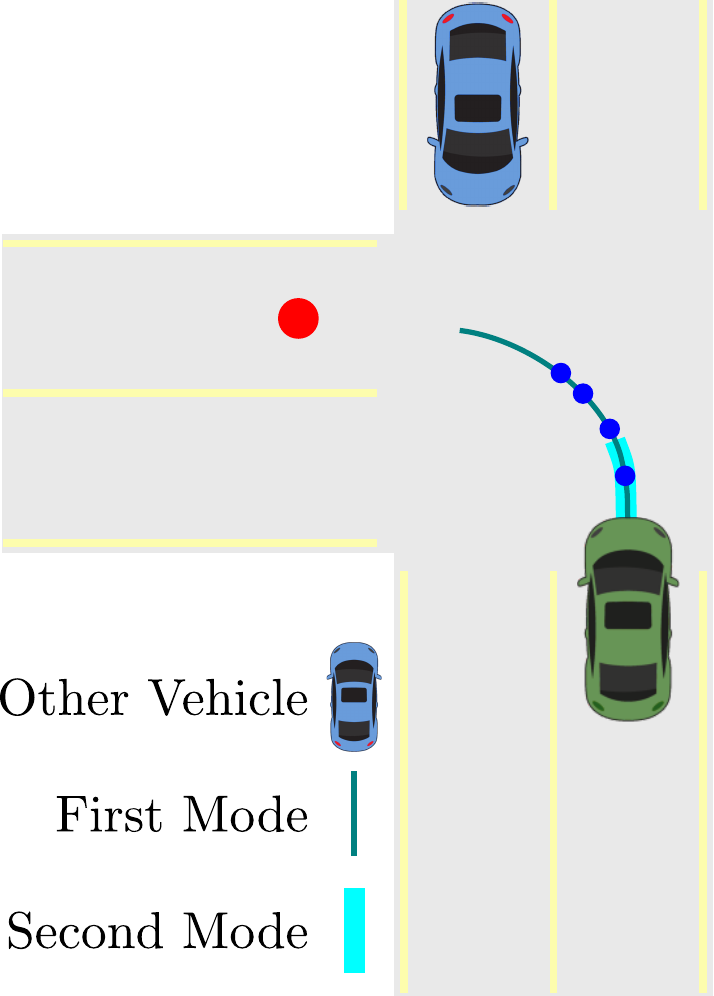}
        \caption{Waypoint ambiguity}
        \label{fig:teaser_2}
    \end{subfigure}
    \vspace{-0.2cm}
   \caption{\textbf{Hidden biases.} (a) When outside their training distribution, current methods extrapolate {\color{blue} waypoint} predictions to the nearest {\color{red}target point}, helping them recover. (b) The future velocity is {\color{mediumteal}multi}-{\color{cyan}modal}, but current methods commit to a single plan, which leads to interpolation.}
\label{fig:teaser}
\vspace{-0.5cm}
\end{figure}

In particular, recent methods trained with Imitation Learning (IL) have shown strong performance~\cite{Chitta2022PAMI, Chen2022CVPRa, Wu2022NeurIPS, Shao2022CORL, Casas2021CVPR}. They are trained using offline datasets, yet they can surprisingly recover from the classic compounding error problem of IL~\cite{Ross2011AISTATS, Prakash2020CVPR}, as indicated by their high route completions~\cite{Chitta2022PAMI, Chen2022CVPRa, Wu2022NeurIPS, Shao2022CORL}. While they do not utilize HD maps as input, they are provided with map-based GNSS locations in the center of the lane (spaced 30 m apart on average) called {target points} (TPs) that describe the route the car should follow. TPs were introduced as an alternative form of conditioning signals to convey driver intent~\cite{Chen2019CORL, Prakash2021CVPR}. Prior work~\cite{Codevilla2018ICRA, Codevilla2019ICCV} used discrete navigation commands or NCs (\ie follow lane, turn right, ...) instead.

In this paper, we show that TP conditioned models recover from the compounding error problem because they use geometric information contained in the TP to reset steering errors periodically (at every TP). This makes them implicitly rely on accurate map information, even though they are otherwise HD map free. Steering directly towards a TP is a shortcut~\cite{Geirhos2020NatureMI} that these IL methods learn to exploit. When methods accumulate enough steering error to be out of distribution during deployment, we observe that they steer towards the nearest TP. When the TP is close, this has the effect of driving back to the lane center, where it is in distribution again. This is illustrated in \figref{fig:teaser_1}. However, when the TP is far away, this shortcut can lead to catastrophic steering errors (\eg cutting a turn). We show examples of this behavior for various SotA architectures in \secref{sec:tp_shortcut}. We demonstrate that the shortcut problem is intrinsically related to the decoder architecture and that a transformer decoder~\cite{Vaswani2017NIPS} can mitigate it. 

Another common aspect of the current SotA is that they use waypoints (future positions of an expert driver) as output representations~\cite{Chitta2022PAMI, Chen2022CVPRa, Wu2022NeurIPS}. We point out that this is an ambiguous representation as the future velocity is multi-modal, yet the model commits to a point estimate. This is illustrated in \figref{fig:teaser_2}. We show that this ambiguity can sometimes be helpful due to the continuous nature of waypoints: the network can continuously interpolate between modes. We propose an alternative that explicitly predicts the uncertainty of the network via target speed classification, and show that interpolating between target speeds weighted by the uncertainty reduces collisions. Our controlled experiments also cover important but sometimes neglected details in the training of end-to-end driving systems, such as augmentation, training schedules, and dataset size. In particular, we revisit the idea of shift and rotation augmentations to aid recovery~\cite{Bojarski2016ARXIV,Amini2022ICRA}. These were common in early IL methods for CARLA with control outputs~\cite{Codevilla2019ICCV}, but are harder to implement with waypoint outputs and not used by the current SotA. We find that they yield significant improvements.

Using these insights, we develop TransFuser++ (TF++), which sets a new SotA on the Longest6~\cite{Chitta2022PAMI} and LAV~\cite{Chen2022CVPRa} benchmarks. Two of the ideas we apply, transformer decoder pooling and path-based outputs instead of waypoints, have been used in Interfuser~\cite{Shao2022CORL}. However, these are presented as minor details, and their impact is not studied in isolation as in our work. TF++ is significantly simpler than Interfuser, yet outperforms it by a large margin on Longest6, as we show in \secref{sec:results}. We use $\sim$4$\times$ less data, 1 camera instead of 4, and do not require complex heuristics to extract throttle and brake commands for our path output. 

\boldparagraph{Contributions}
\begin{itemize}[noitemsep,topsep=0pt]
    \item We show that target point conditioned models learn a shortcut that helps them recover from steering errors.
    \item We point out that the waypoint output representation is ambiguous, but its continuous nature helps models collide less by interpolating to slow down.
    \item Using the insights gained from controlled experiments, we propose TransFuser++ which places first on the Longest6 and LAV benchmarks.
\end{itemize}

\noindent Code, data, and models are available at \url{https://github.com/autonomousvision/carla_garage}.
\section{Related Work}
\label{sec:related}

IL for autonomous driving dates back over 30 years~\cite{Pomerleau1988NIPS}. It regained traction with the seminal works of~\cite{Bojarski2016ARXIV, Codevilla2018ICRA, Bansal2019RSS} and the release of CARLA~\cite{Dosovitskiy2017CORL}, a 3D simulator used in numerous recent research advances in autonomous driving~\cite{Sauer2018CORL, Codevilla2019ICCV, Chen2019CORL, Toromanoff2020CVPR, Ohn-Bar2020CVPR, Chitta2021ICCV, Zhang2021ICCV, Hu2022NeurIPS}. Early IL approaches evaluated in CARLA used a discrete navigation command (NC)~\cite{Codevilla2019ICCV, Chen2019CORL, Chen2021ICCVa}, but their performance is not competitive to modern approaches~\cite{Chitta2022PAMI, Chen2022CVPRa, Wu2022NeurIPS, Shao2022CORL} which predict waypoints conditioned on TPs. In this work, we revisit an idea used in early systems but neglected in modern TP-conditioned methods: geometric shift augmentations~\cite{Bojarski2016ARXIV,Chen2019CORL,Codevilla2019ICCV,Prakash2020CVPR}.

\textbf{LAV}~\cite{Chen2022CVPRa} supervises waypoint outputs with additional data by making predictions for other nearby agents in the scene. Their waypoints are initially generated with NC conditioning. These are then refined using a GRU \cite{Cho2014Proc.oftheConferenceonEmpiricalMethodsinNaturalLanguageProcessingEMNLP}. They observe a large (+50) improvement in route completion from this refinement. Our findings suggest that the refinement module improves steering primarily through TP conditioning, which is only provided to the refinement GRU.
\textbf{TCP}~\cite{Wu2022NeurIPS} observes that waypoints are stronger at collision avoidance than directly predicting controls, but sometimes suffer at large turns. They leverage the strength and mitigate the weaknesses of these two representations by creating a situation dependent ensemble. Our study suggests that the waypoints are better at collision avoidance due to an implicit slowdown when the car is uncertain.
\textbf{PlanT}~\cite{Renz2022CORL} investigates planning on CARLA by processing object bounding boxes with a transformer. Its sensorimotor version differs from other SotA models in that it is trained in two stages (not end-to-end). In our setting, we find that end-to-end training is crucial. We show that their observations about dataset scale also hold for end-to-end models. 

\textbf{TransFuser} \cite{Chitta2022PAMI, Prakash2021CVPR} is a simple, well-known and widely used baseline for CARLA. We provide an explanation for the large differences in route completion between the target point conditioned TransFuser architecture and its NC conditioned baselines~\cite{Codevilla2019ICCV,Chen2019CORL,Chen2021ICCVa}. Further, we propose modifications to its architecture, output representation and training strategy which lead to significant improvements. 

\textbf{Interfuser} \cite{Shao2022CORL} regresses a path for steering, predicts object density maps, and classifies traffic rule flags. This representation is converted by a forecasting mechanism and hand designed heuristics into control. Like our proposed method, it uses a transformer decoder for pooling features, and disentangles future velocities from the path in the output. While these ideas were already present, they were not studied or discussed as significant. Our work adds to the literature by showing that these design choices are indeed critical to performance and providing explanations as to why. Furthermore, our final system is simpler and significantly outperforms Interfuser on Longest6.

\textbf{ReasonNet} \cite{Shao2023CVPR}, \textbf{ThinkTwice} \cite{Jia2023CVPR} and \textbf{CaT} \cite{Zhang2023CVPR} are concurrent end-to-end driving approaches that all use TP conditioning and waypoints as output representation.

\begin{figure*}
\centering
    \begin{subfigure}[b]{0.33\textwidth}
        \centering
        \includegraphics[width=\textwidth]{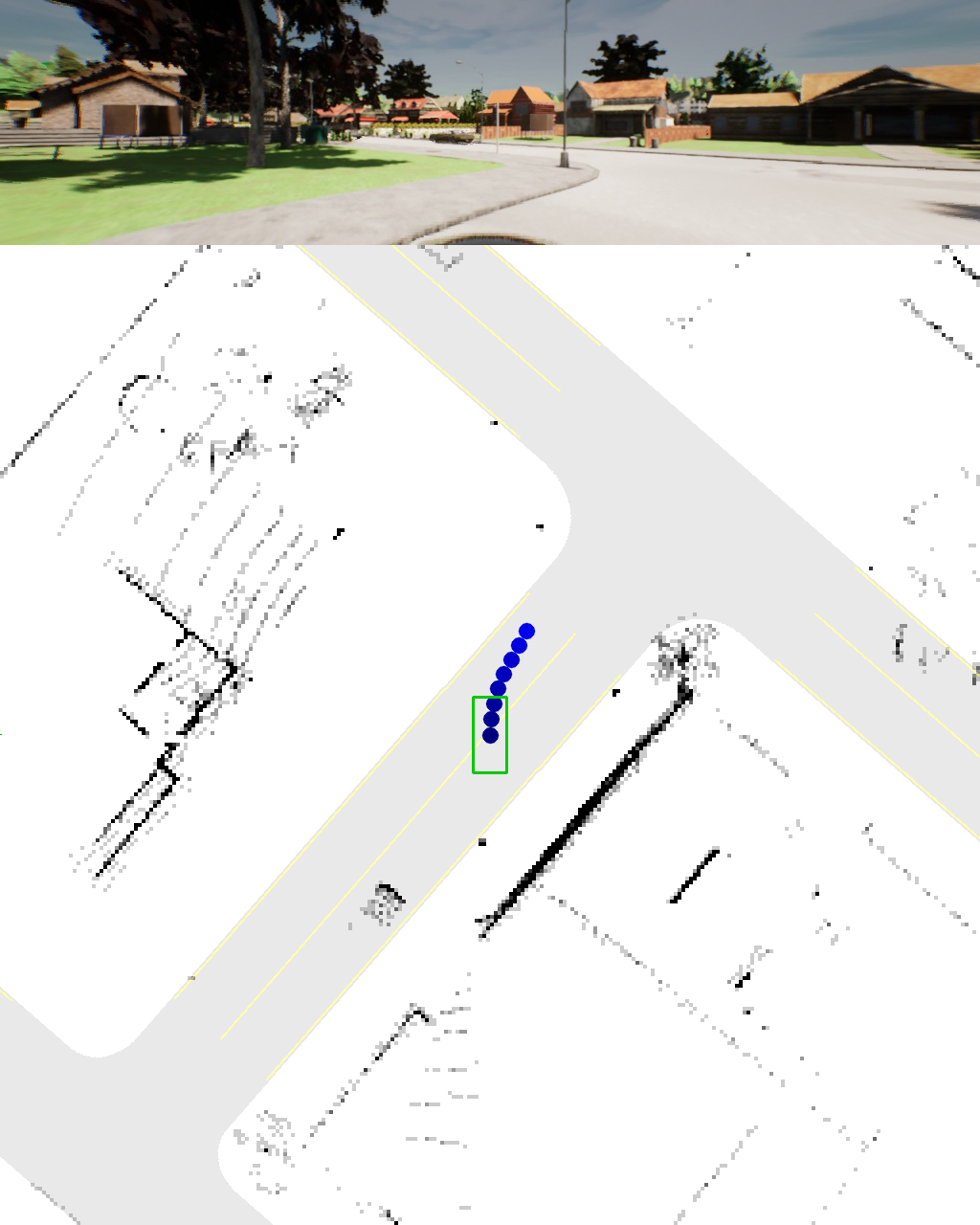}
        \caption{TransFuser (NC conditioned)}
        \label{fig:discrete_target_point_failure}
    \end{subfigure}
    \begin{subfigure}[b]{0.33\textwidth}  
        \centering 
        \includegraphics[width=\textwidth]{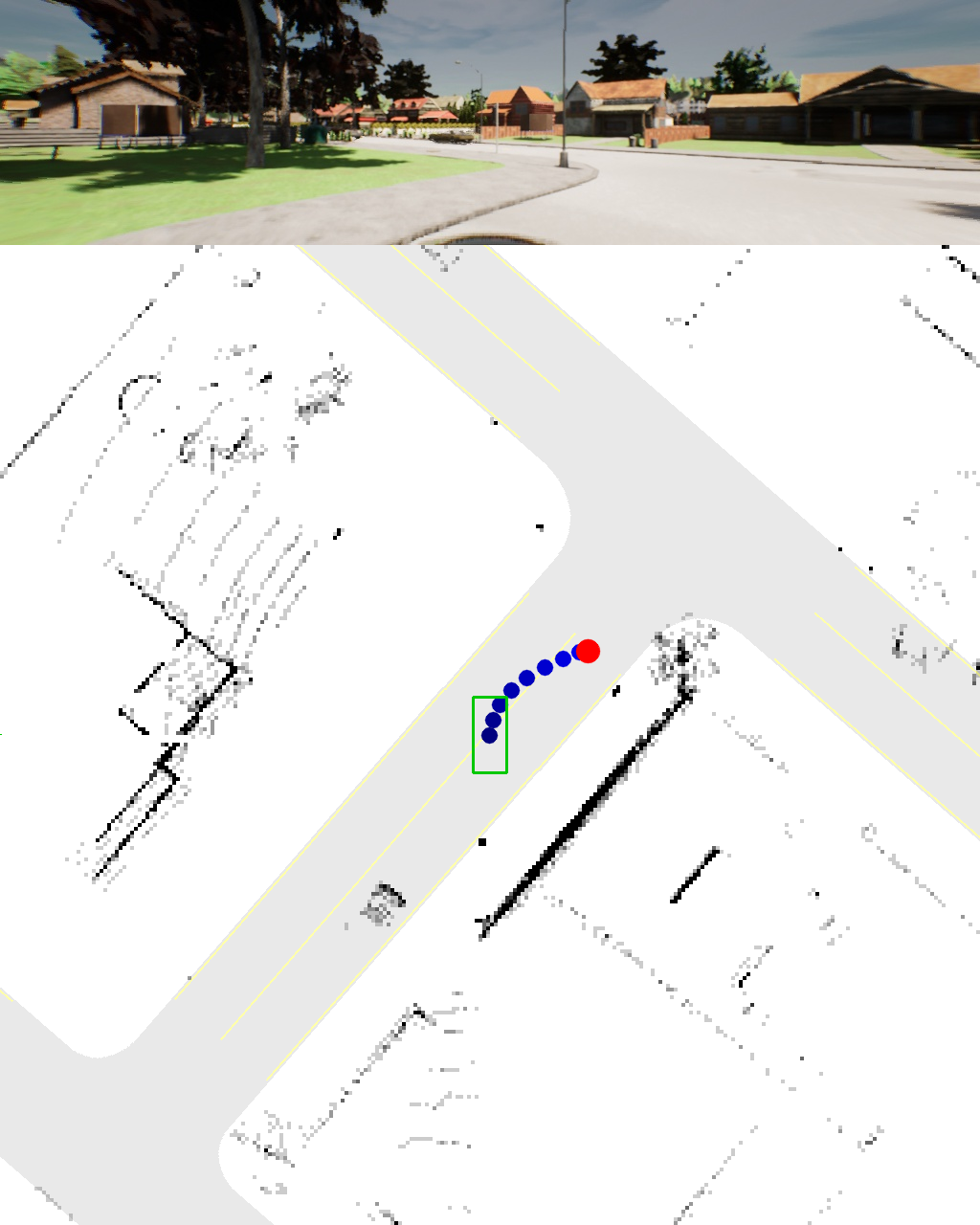}
        \caption{TransFuser (TP conditioned)}
        \label{fig:target_point_success}
    \end{subfigure}
    \begin{subfigure}[b]{0.33\textwidth}
        \centering
        \includegraphics[width=\textwidth]{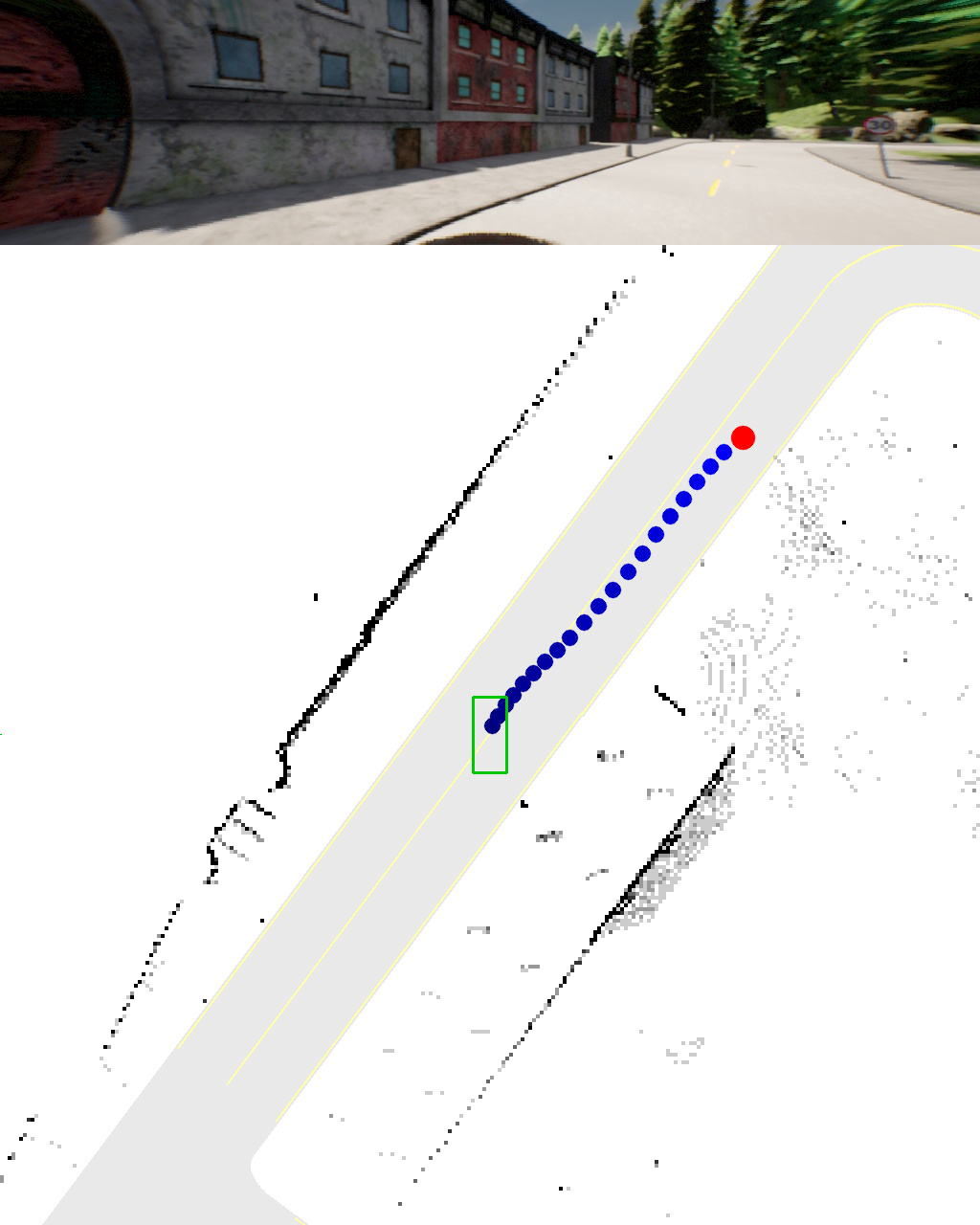}
        \caption{LAV~\cite{Chen2022CVPRa} (TP conditioned)}
        \label{fig:target_point_success_lav}
    \end{subfigure}
    \vspace{-0.7cm}
   \caption{\textbf{Extrapolation to target point.} In unknown situations, TP conditioned methods extrapolate their \textcolor{blue}{waypoints} towards \textcolor{red}{target points}. This periodically resets steering errors and is a form of implicit map based recovery. However, relying on extrapolation is a shortcut that can lead to catastrophic errors in certain situations (see \figref{fig:target_point_failure_tcp} and \figref{fig:target_point_failure}).}
\label{fig:TP_Recovery_OOD}
\vspace{-0.3cm}
\end{figure*}

\section{Hidden Biases of End-to-End Driving}
\label{sec:secrets}

We consider the task of urban navigation from point A to B~\cite{Chitta2022PAMI}. The goal is to complete routes with dense traffic, multiple lanes and complex geometries (\eg roundabouts) without incurring infractions. Along the way, the agent encounters manually designed pre-crash traffic scenarios. Each route is a list of GNSS coordinates called target points (TPs) which can be up to 50 m apart ($\sim$30m on average).

\boldparagraph{Metrics} We use the CARLA online metrics. Route Completion (RC) is the percentage of the route completed. Infraction Score (IS) is a penalty factor starting at 1.0 that gets reduced multiplicatively for every infraction. Our main metric is the Driving Score (DS) which multiplies RC with the IS. Where insightful, we report infraction per kilometer metrics. For a comprehensive description, refer to \cite{Chitta2022PAMI}.

\boldparagraph{Baseline} As a baseline, we reproduce TransFuser~\cite{Chitta2022PAMI}. This is a simple method representative of current SotA driving systems on CARLA. The main differences in our reproduction are a 360\textdegree\ field of view (FOV) LiDAR to enable safe lane changes and a dataset where the expert is driving 2$\times$ faster to speed up evaluation. A list of other minor changes is provided in the supplementary material.

\boldparagraph{Benchmark} For all experiments in this section, we train on CARLA towns 01, 03, 04, 06, 07 and 10. We use the 16 validation routes from LAV~\cite{Chen2022CVPRa} in the withheld towns 02 and 05 for evaluation. In order to reduce the influence of training and evaluation variance which can be large in CARLA~\cite{Behl2020IROS, Chitta2022PAMI}, every experiment reports the average of 3 training seeds, evaluated 3 times each. We additionally report the training standard deviation for the main metrics.

\subsection{A shortcut for recovery}
\label{sec:tp_shortcut}
While current SotA CARLA methods have fundamentally different architectures~\cite{Chitta2022PAMI, Shao2022CORL, Wu2022NeurIPS, Chen2022CVPRa}, they are all trained with conditional IL using fixed pre-recorded datasets. The evaluation task involves challenging routes which are $\sim$1.5km long, so it would be expected that these methods suffer from \textit{compounding errors}, a well-known problem for IL~\cite{Ross2011AISTATS}. Geometric shift augmentations~\cite{Bojarski2016ARXIV} are a common approach to teach IL methods how to recover from such compounding steering errors~\cite{Bojarski2016ARXIV,Codevilla2019ICCV}. Surprisingly, even though such augmentations are not employed by SotA methods on CARLA, they report high RCs, demonstrating that they are not prone to this expected failure mode.

All the aforementioned methods condition their predictions using the next target point (TP) along the route. To understand the importance of the TP, we train 2 versions of our reproduction of TransFuser: one with the original TP conditioning and another with a discrete (NC) condition. This is a one hot vector that indicates whether the car should follow the lane, turn right at the next intersection, etc. It contains no geometric information about the center of the lanes, but still removes the inherent task ambiguity of inner-city driving~\cite{Codevilla2018ICRA}. For the NC conditioned model, we also implement shift and rotation augmentations by collecting augmented frames during data collection. We deploy a second camera in the simulator, that changes its position and orientation randomly at every time-step. The output labels are transformed accordingly for training the model (implementation details and examples can be found in the supplementary material). The results are shown in \tabref{tab:discrete_command}. 

\begin{table}[h]
\small
\centering
    \begin{tabular}{l | c | c c | c }
        \toprule
        \textbf{Cond.} & \textbf{Aug.} & \textbf{DS} $\uparrow$ & \textbf{RC} $\uparrow$ & \textbf{Dev} $\downarrow$ \\
        \midrule
        NC & {-} & {32} \pmsd {8} & 56 \pmsd {12} & {0.86} \\
        NC & {$\checkmark$} & {35} \pmsd {3} & 54 \pmsd {4} & {0.99} \\
        TP & {-} & \textbf{39} \pmsd {9} & \textbf{84} \pmsd {7} & \textbf{0.00}\\
        \bottomrule
    \end{tabular}
    \caption{\textbf{Conditioning and Augmentation.}}
    \label{tab:discrete_command}
    \vspace{-0.3cm}
\end{table}

\begin{figure*}
\centering
    \begin{subfigure}[b]{0.33\textwidth}
        \centering
        \includegraphics[width=\textwidth]{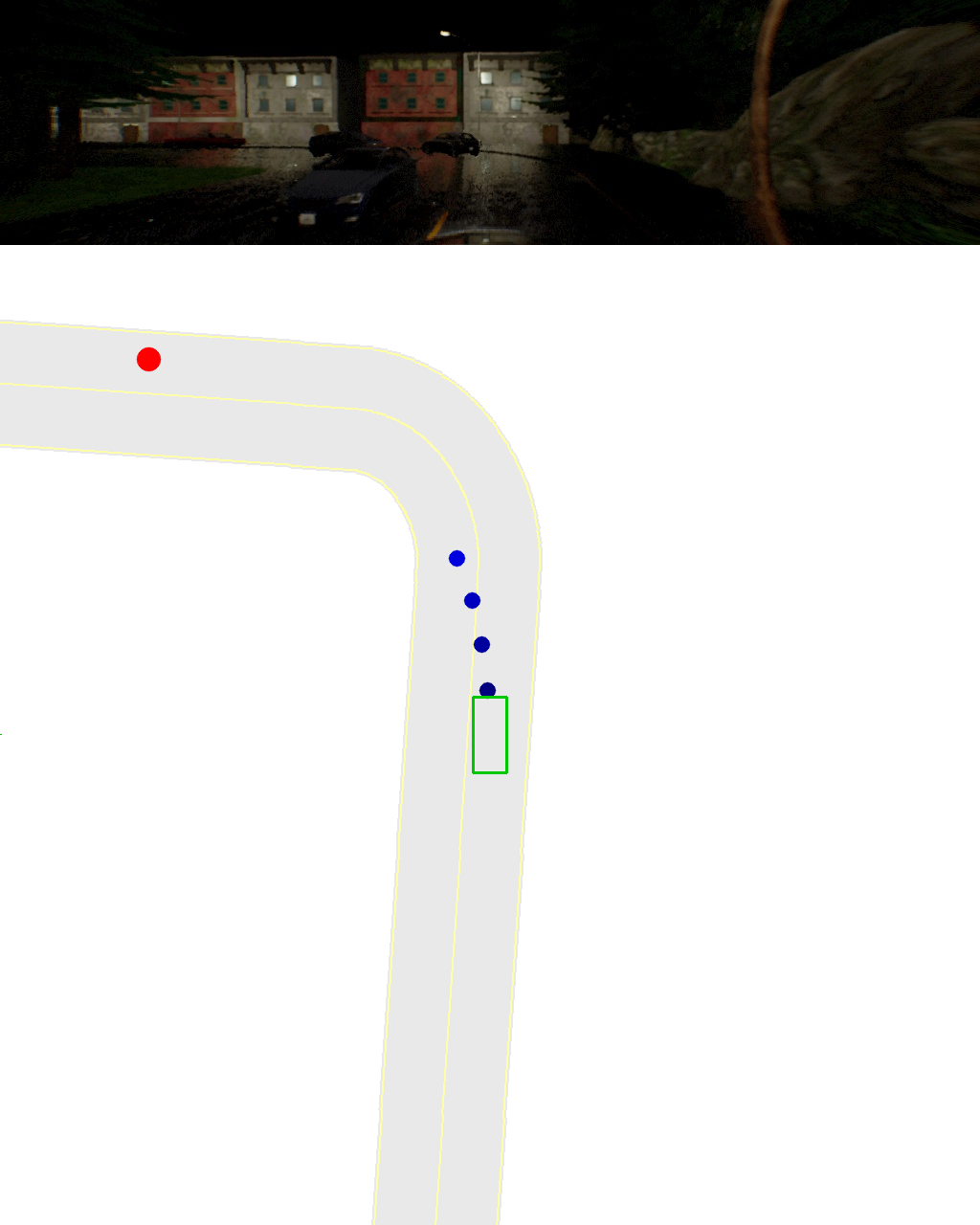}
        \caption{TCP~\cite{Wu2022NeurIPS} follows a shortcut.}
        \label{fig:target_point_failure_tcp}
    \end{subfigure}
    \begin{subfigure}[b]{0.33\textwidth}
        \centering
        \includegraphics[width=\textwidth]{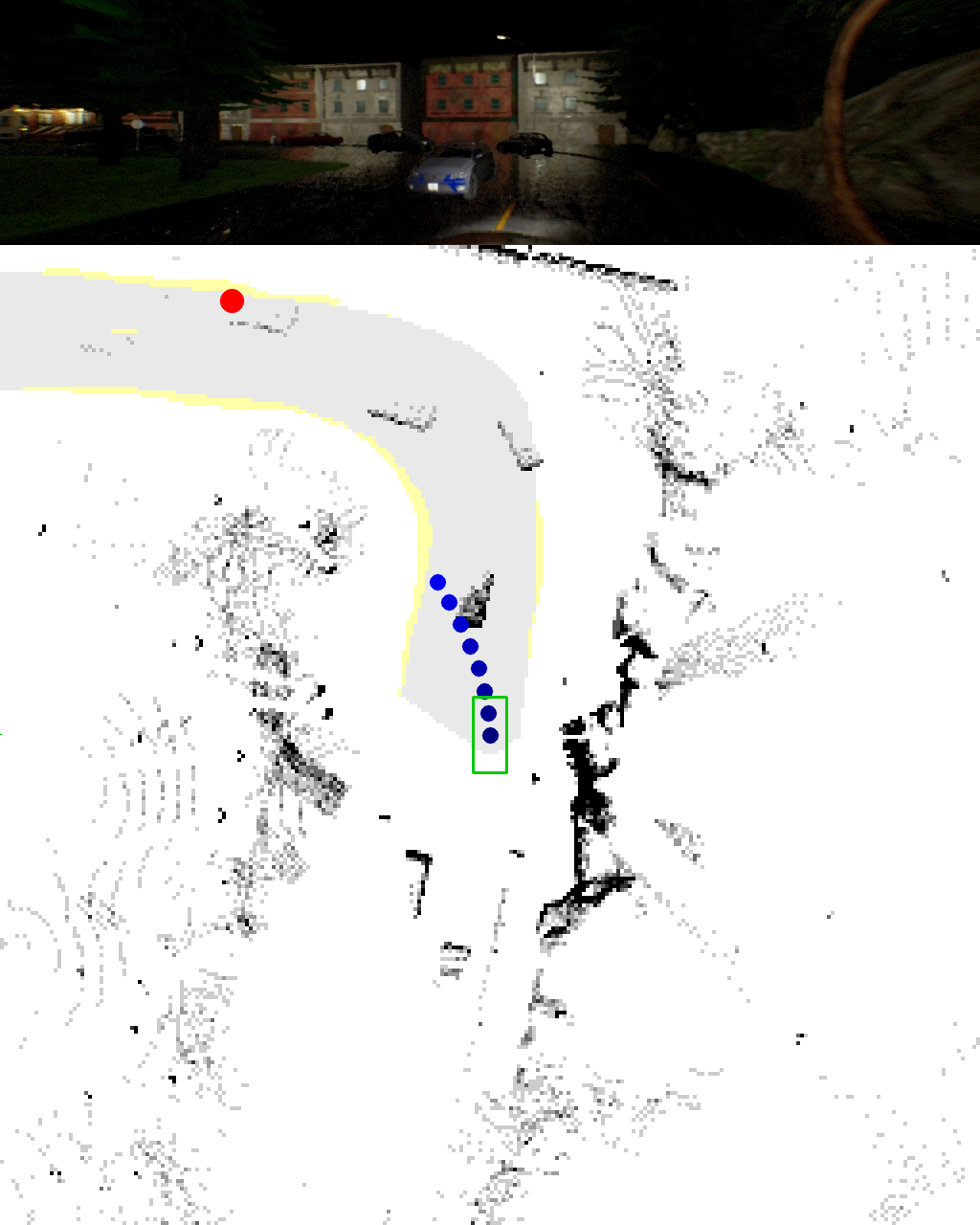}
        \caption{TransFuser follows a shortcut.}
        \label{fig:target_point_failure}
    \end{subfigure}
    \begin{subfigure}[b]{0.33\textwidth}  
        \centering 
        \includegraphics[width=\textwidth]{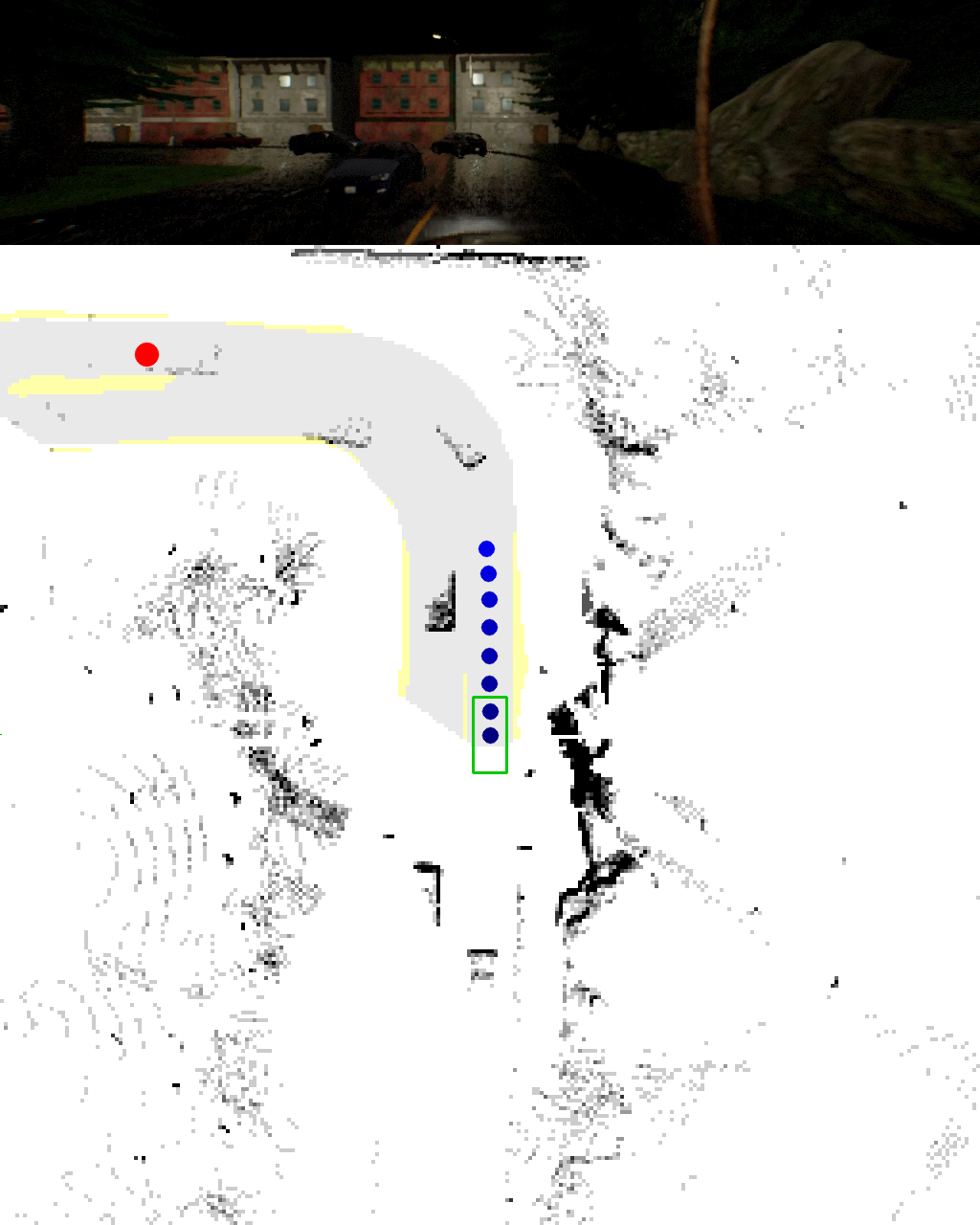}
        \caption{A transformer decoder fixes this.}
        \label{fig:target_point_success_2}
    \end{subfigure}
    \vspace{-0.7cm}
   \caption{\textbf{Target point shortcut.} When TP conditioned methods extrapolate to spatially distant waypoints, they incur large steering errors. Replacing global average pooling in TransFuser with a cross-attention mechanism mitigates the issue.}
\label{fig:target_point}
\vspace{-0.3cm}
\end{figure*}

We observe a significant difference in RC of 28 points when switching between TP and NC conditioning. The TP conditioned model has 0 route deviations per km (``Dev"), which indicates that it never takes a wrong turn or drives too far away (more than 30m) from the lane. However, this is not the case for NC conditioned models. While we do see a small improvement in DS with augmentation, its RC is still unsatisfactory. This indicates that the geometric information regarding the lane center available with the TP aids recovery in SotA IL methods, prompting further investigation. We inspect the TransFuser and LAV~\cite{Chen2022CVPRa} models by constructing situations where the car is forcefully steered out of lane. \figref{fig:TP_Recovery_OOD} visualizes these scenarios with the camera, LiDAR and ground truth HD maps. For TP conditioned models, the predicted waypoints (shown in blue) extrapolate towards the nearest TP (red) even though the situation is out of distribution. This shows that one reason for their strong route following is that the bird's eye view (BEV) TP resets steering errors periodically: methods learn to steer towards nearby TPs because the expert trajectory in the dataset always goes through them. This has the effect that TP conditioned methods periodically drive back to the center of the lane, resetting any accumulated errors. Furthermore, it suggests that SotA IL approaches strongly rely on pre-specified geometric information regarding lane centers for recovery.

We identify this as a form of shortcut learning, which can be useful when the car is close to a target point, but can also lead to catastrophic steering errors when the target point is further away. An example would be directly extrapolating to a target point behind a turn, which leads to cutting the turn. In \figref{fig:target_point_failure_tcp} and \figref{fig:target_point_failure}, we show instances where this happens for TCP~\cite{Wu2022NeurIPS} and TransFuser~\cite{Chitta2022PAMI} in a validation town at nighttime. TransFuser also predicts the BEV segmentation as an auxiliary task, which we overlay in the figure (gray: route, yellow: lane marking). For TCP, we render the ground truth map and omit the LiDAR, since it does not use LiDAR and does not predict BEV segmentation. Both methods predict waypoints that are tilted towards the TP instead of following the street. As a result, they drive into the opposing lane. We refer to this as the \textit{TP shortcut}.

\begin{figure}[t]
\begin{center}
   \includegraphics[width=\linewidth]{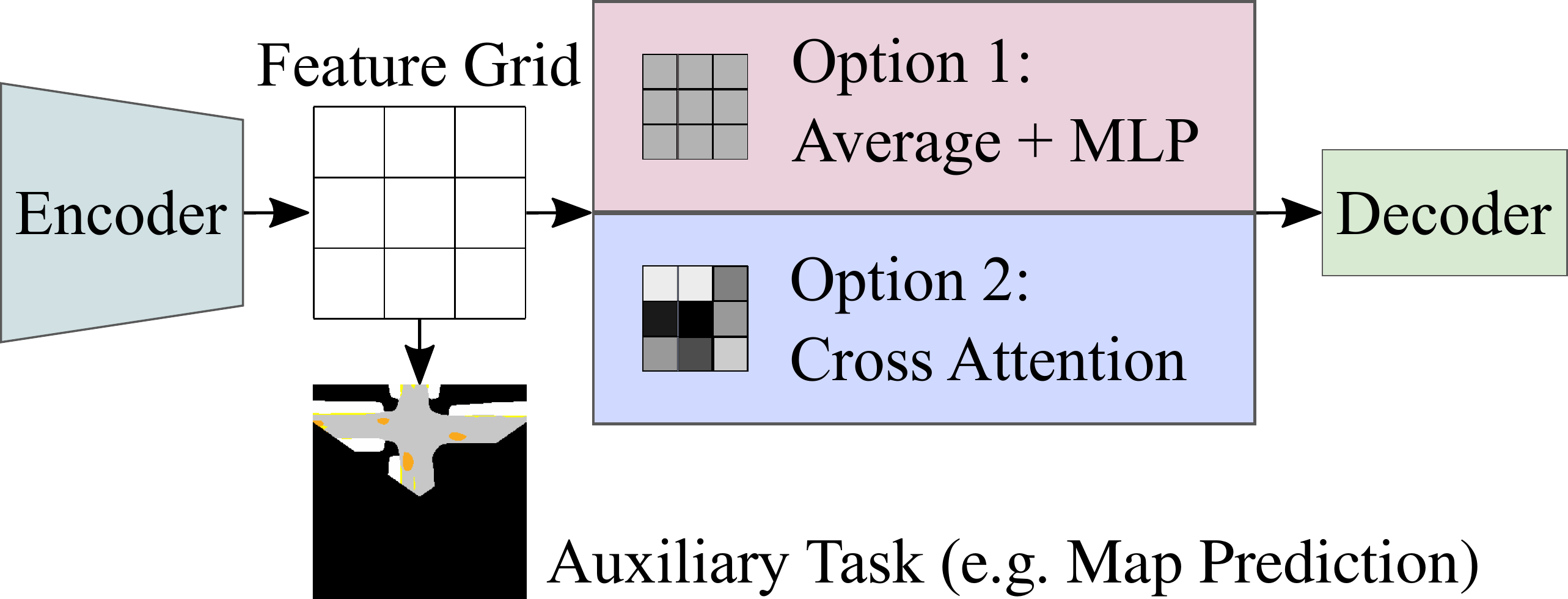}
\end{center}
\vspace{-0.3cm}
\caption{\textbf{Pooling.} Existing approaches vectorize feature grids either by global average pooling (top, \eg~\cite{Chitta2022PAMI,Chen2022CVPRa}) or with attention mechanisms (bottom, \eg~\cite{Shao2022CORL,Wu2022NeurIPS}). The latter retains spatial information gained via auxiliary tasks.}
\label{fig:pooling}
\vspace{-0.3cm}
\end{figure}

\subsection{Improved pooling and data augmentation}
\label{sec:mitigating_tp}

One design choice which differs across SotA architectures is the pooling applied between the encoder and decoder. In \figref{fig:pooling}, we summarize these. In particular, TransFuser~\cite{Chitta2022PAMI} and LAV~\cite{Chen2022CVPRa} employ global average pooling (GAP) followed by an MLP. InterFuser~\cite{Shao2022CORL} pools features via the cross-attention mechanism of a transformer decoder. Finally, TCP~\cite{Wu2022NeurIPS}, which has two decoders, uses GAP for one decoder and attention-based pooling for the other.

As shown in \figref{fig:pooling}, the networks are encouraged to learn spatially meaningful feature grids through auxiliary convolutional decoders that predict outputs such as BEV semantic segmentation. However, GAP does not maintain the spatial information in the features. \figref{fig:target_point_failure} shows the BEV road segmentation predicted by TransFuser overlaid by its LiDAR input. Unlike the waypoints, the BEV predictions are quite accurate. This indicates strong features coming from the encoder. Nevertheless, the final conditional waypoint predictions focus on the TP signal in this situation. We hypothesize that the GAP operation makes it difficult for the downstream decoder to utilize the strong BEV features. Note that attention-based pooling, such as the transformer decoder of Interfuser, preserves spatial information through the use of positional encodings.

We compare the original TransFuser GAP approach with the spatially preserving design of Interfuser in \tabref{tab:feature_pooling}. For the latter variant, we remove the GAP operation and MLP in TransFuser. We then process the 8$\times$8 BEV feature grid as tokens with a transformer decoder. Implementation details can be found in the supplementary material. 

\begin{table}[h]
\small
\centering
    \begin{tabular}{l | c | c c | c}
        \toprule
        \textbf{Pooling} & \textbf{Aug.} & \textbf{DS} $\uparrow$ & \textbf{RC} $\uparrow$ & \textbf{Stat} $\downarrow$ \\
        \midrule
        GAP + MLP & - & {39} \pmsd {9} & {84} \pmsd {7} & {1.04}\\
        Transformer Decoder & -  & {43} \pmsd {6} & \textbf{93} \pmsd {3} & {0.55} \\
        Transformer Decoder & $\checkmark$ & \textbf{49} \pmsd {8} & {90} \pmsd {4} & \textbf{0.10} \\
        \bottomrule
    \end{tabular}
    \caption{\textbf{Pooling and augmentation.}}
    \label{tab:feature_pooling}
    \vspace{-0.3cm}
\end{table}

After replacing GAP with the transformer decoder, the RC increases by 9 points and collisions per km with static objects (``Stat") reduce by a factor of 2. Static objects (such as poles) only occur outside lanes in CARLA, implying that the network manages to stay in the lane far more consistently. \figref{fig:target_point_success_2} provides qualitative evidence that the transformer decoder can succeed in situations where GAP failed due to the TP shortcut (additional examples in supplementary). In addition, we investigate the inclusion of shift and rotation augmentations designed to aid lateral recovery (as described in \secref{sec:tp_shortcut}). Collisions with the static environment are reduced further by a factor of 5 indicating that augmentation provides strong benefits.

\subsection{The ambiguity of waypoints}
\label{sec:disentanglement}

Waypoints are used in many SotA systems as outputs~\cite{Chen2022CVPRa, Wu2022NeurIPS, Chitta2022PAMI}. They are obtained by recording an expert driver's GNSS locations at fixed time intervals (\eg 0.5s) and transforming them into a local coordinate frame. The model is then trained to predict future waypoints, typically with an $L_1$ regression objective. Waypoints entangle both the path and the future velocities of the vehicle. The velocity after a specific time interval extracted from the waypoints is used by a downstream controller as a target speed. 

\begin{figure}[t]
\vspace{-0.7cm}
\begin{center}
   \includegraphics[width=\linewidth]{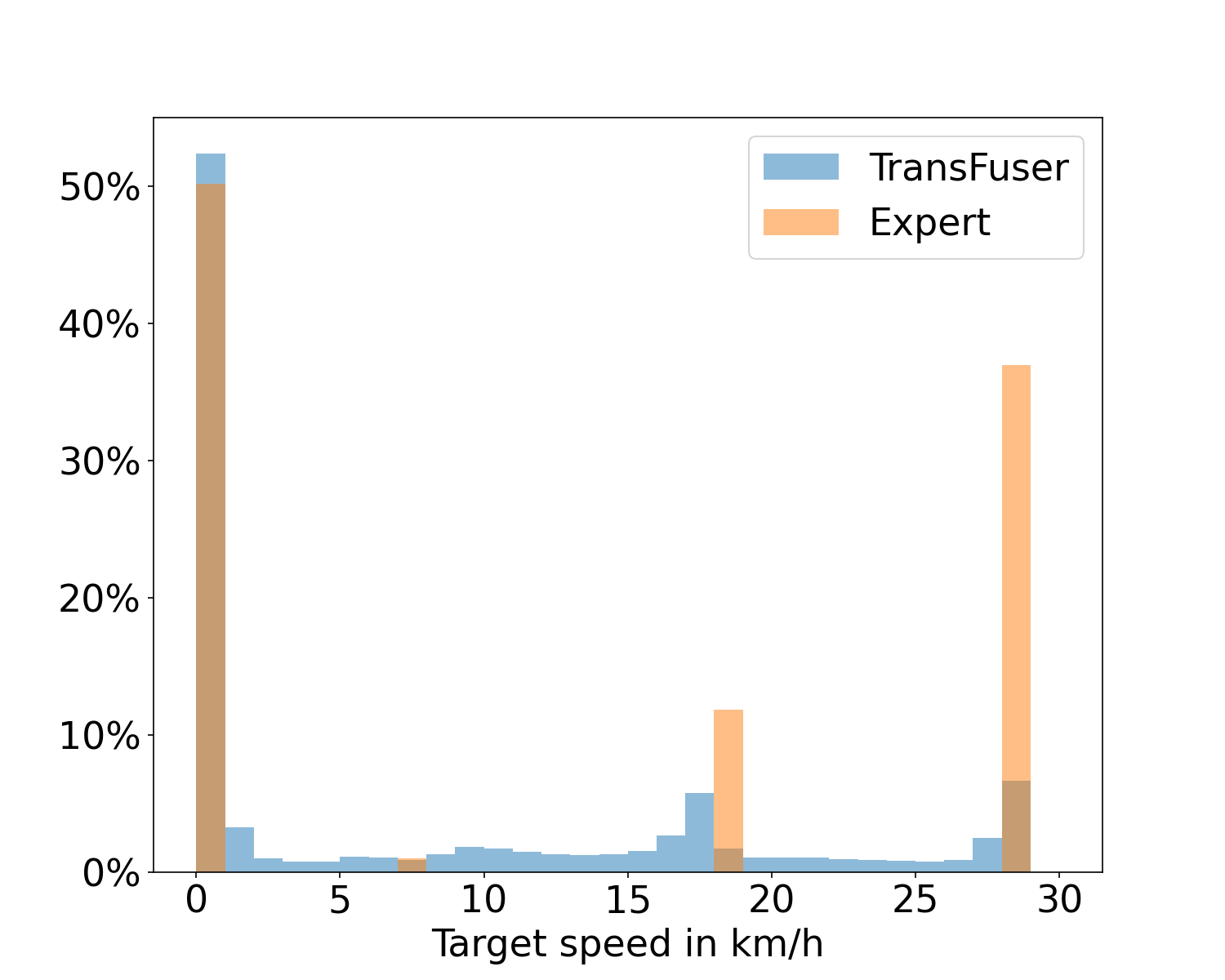}
\end{center}
\vspace{-0.5cm}
\caption{\textbf{Transfuser interpolates between modes.}}
\label{fig:histogram}
\vspace{-0.5cm}
\end{figure}

In \figref{fig:histogram}, we plot a histogram comparing the target speeds of the expert algorithm (\ie training dataset) to those extracted from TransFuser (best model in \tabref{tab:feature_pooling}). Interestingly, the distributions are quite different. By design, the expert chooses one of four target speeds: 29, 18, 7, or 0 km/h. These values cover behaviors needed for city driving, corresponding to four situations: regular driving, slowing down in intersections, slowing down near pedestrians, and stopping. In contrast, for TransFuser, the predicted target speeds cover the entire 0-29 km/h range.

\begin{figure*}[t]
\centering
    \begin{subfigure}[b]{0.33\textwidth}
        \centering
        \includegraphics[width=\textwidth]{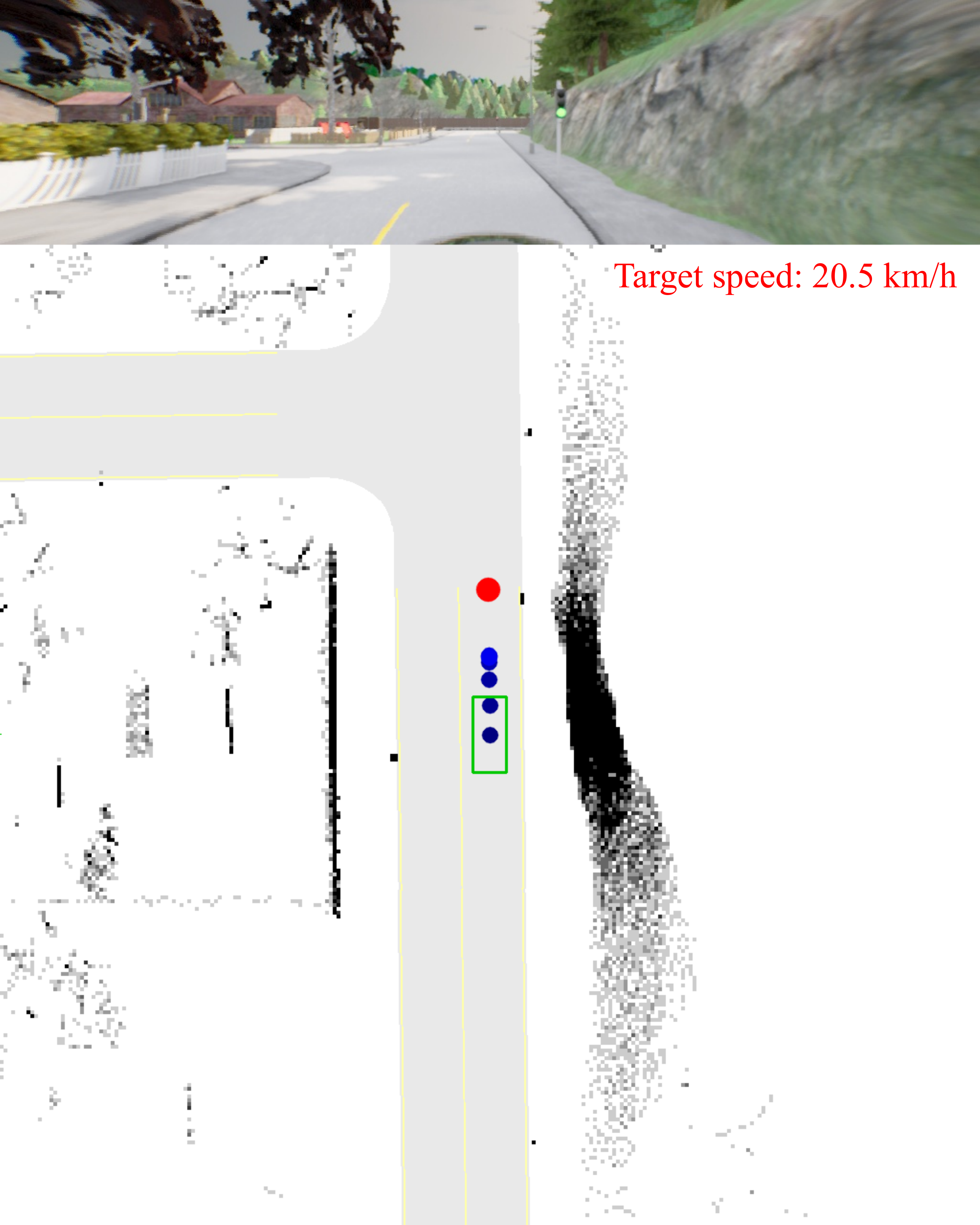}
        \caption{Slowing down at a green light.}
        \label{fig:conservative_mode}
    \end{subfigure}
    \begin{subfigure}[b]{0.33\textwidth}  
        \centering 
        \includegraphics[width=\textwidth]{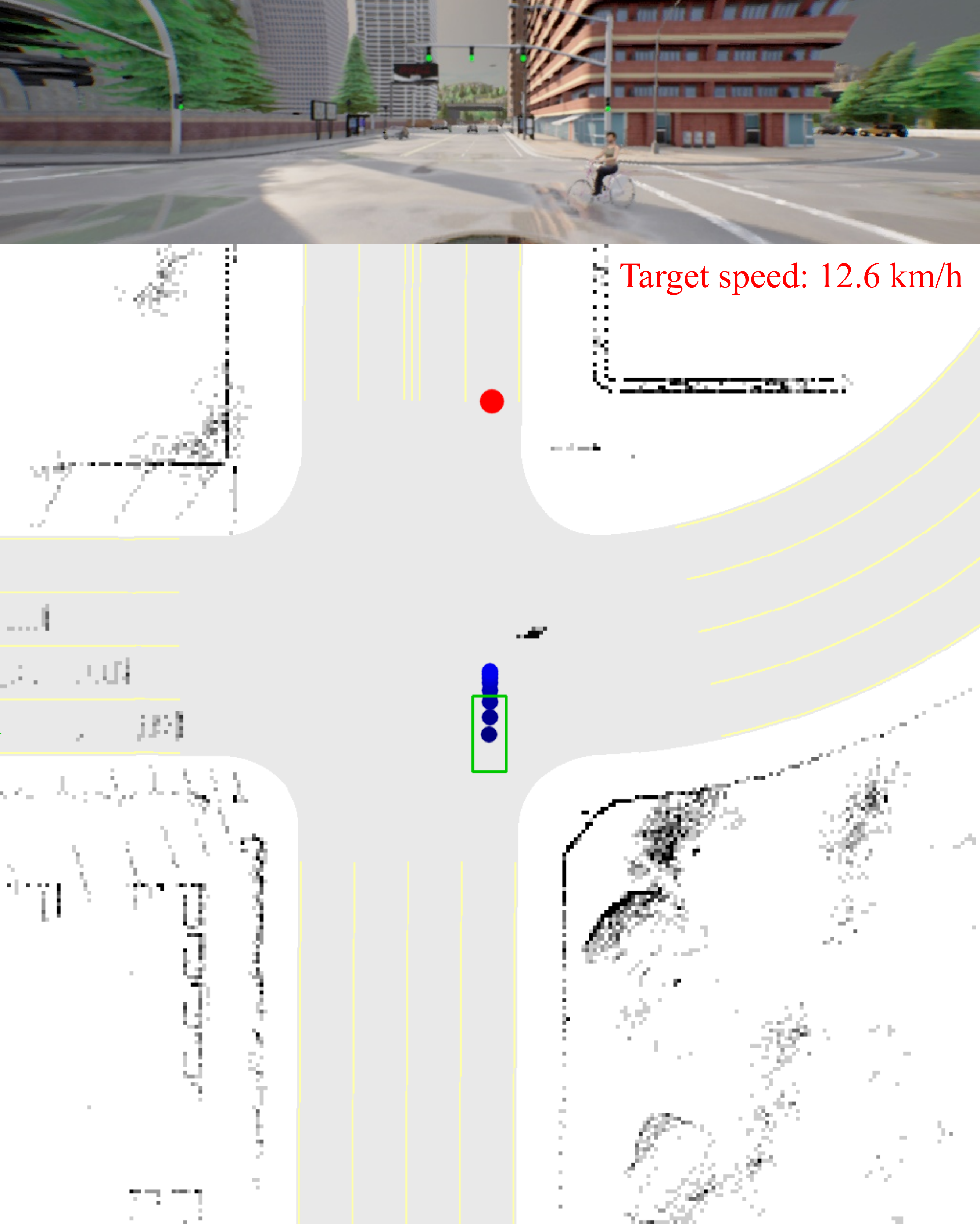}
        \caption{Slowing down in an intersection.}
        \label{fig:aggressive_mode}
    \end{subfigure}
    \begin{subfigure}[b]{0.33\textwidth}  
        \centering 
        \includegraphics[width=\textwidth]{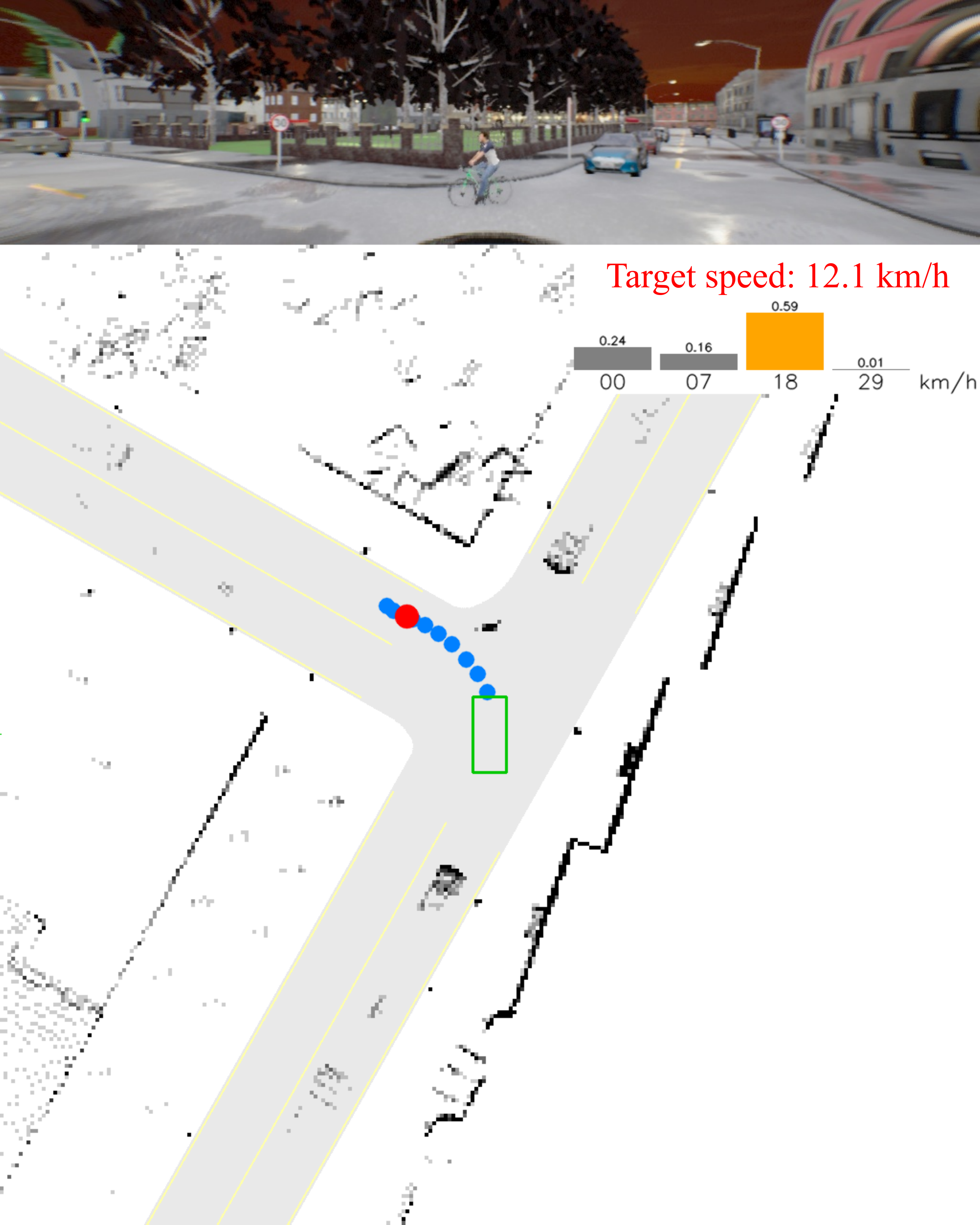}
        \caption{Disentangling route and target speed.}
        \label{fig:interpolating_modes}
    \end{subfigure}
    \vspace{-0.7cm}
   \caption{\textbf{Waypoints are ambiguous.} The model's output representation forces it to predict a single mode for future velocities. There is a possibility that the traffic light might turn red in the future, or that the cyclist either cuts in or yields to the agent. The particular mode (or interpolation thereof) that the model converges to, depends on the training run and dataset.}
\label{fig:waypoint_problem}
\vspace{-0.3cm}
\end{figure*}

While the future path that our expert follows is deterministic and unambiguous (center of the lane, lane change at pre-defined locations), future velocities are multi-modal. As both are jointly represented by waypoints, this leads to an ambiguous entangled representation. However, existing methods which utilize this entangled waypoint representation predict only point estimates (\ie, a single set of waypoints) as output, hence only a single mode is modeled. From \figref{fig:histogram}, we observe that the waypoint based TransFuser model indeed interpolates between modes, a behavior that is expected when modeling multi-modal outputs deterministically. We illustrate this behavior in detail using the examples of (1) approaching an intersection with a green light (\figref{fig:conservative_mode}), and (2) a cyclist cutting into the vehicle's path (\figref{fig:aggressive_mode}). TransFuser slows down in these situations since it is uncertain, \eg, the light may turn red in \figref{fig:conservative_mode}. The car stops in time in \figref{fig:aggressive_mode} because of the decreased speed.

While this averaging is indeed beneficial in some situations, an entangled representation is undesirable as it is less interpretable and does not explicitly expose uncertainty. Moreover, when stopped (all waypoints collapsed to one location), the steering signal is undefined which requires additional heuristics in the controller. To resolve this, we disentangle the future velocities from the path by sampling the expert's position at fixed distances instead of fixed time intervals for training a (deterministic) \textit{path predictor}. However, when predicting the path instead of time-dependent trajectories, one requires an additional method to determine a target speed for the car. We propose to predict the target speed by simple classification using an additional MLP head. We incorporate the prediction uncertainties into the final output by using a confidence weighted average of the predicted target speed as input to the controller. Note that this is just one example of how the resulting confidence estimates may be leveraged by a vehicle controller.

\begin{table}[h]
\small
\centering
    \begin{tabular}{l | c c | c c}
        \toprule
        \textbf{Output} & \textbf{DS} $\uparrow$ & \textbf{RC} $\uparrow$ & \textbf{Veh} $\downarrow$ & \textbf{Stat} $\downarrow$ \\
        \midrule
        Waypoints & {49} \pmsd {8} & \textbf{90} \pmsd {4} & \textbf{0.70} & {0.10} \\
        Path + Argmax &  {40} \pmsd {1} & {88} \pmsd {2} & {1.25} & {0.03} \\
        Path + Weighted &  \textbf{50} \pmsd {3} & {88} \pmsd {1} & {0.83} & \textbf{0.02} \\
        \bottomrule
    \end{tabular}
    \caption{\textbf{Output representation.}}
    \label{tab:disentanglement}
    \vspace{-0.5cm}
\end{table}

Compared to the waypoint representation, naively converting the predicted classes to a target speed via the most likely class (Argmax) increases vehicle collisions (``Veh"), as seen in \tabref{tab:disentanglement}. Employing the proposed weighting (Weighted) achieves lower collisions. Moreover, the path-based model steers better as indicated by the lower static obstacle collisions (``Stat"). The disentangled representation achieves the same driving score as the entangled waypoint representation, providing an alternative with a simpler and more interpretable controller: (1) it eases design since it has identical parameters to the controller in the expert, (2) it has access to an unambiguous path representation even when driving slowly, and (3) it explicitly exposes and makes use of uncertainties with regard to target speeds.

\figref{fig:interpolating_modes} shows a qualitative example. The car is slowing down due to its uncertainty. The lower target speed allows it successfully merge behind the cyclist.

\begin{figure}[t]
\begin{center}
   \includegraphics[width=\linewidth]{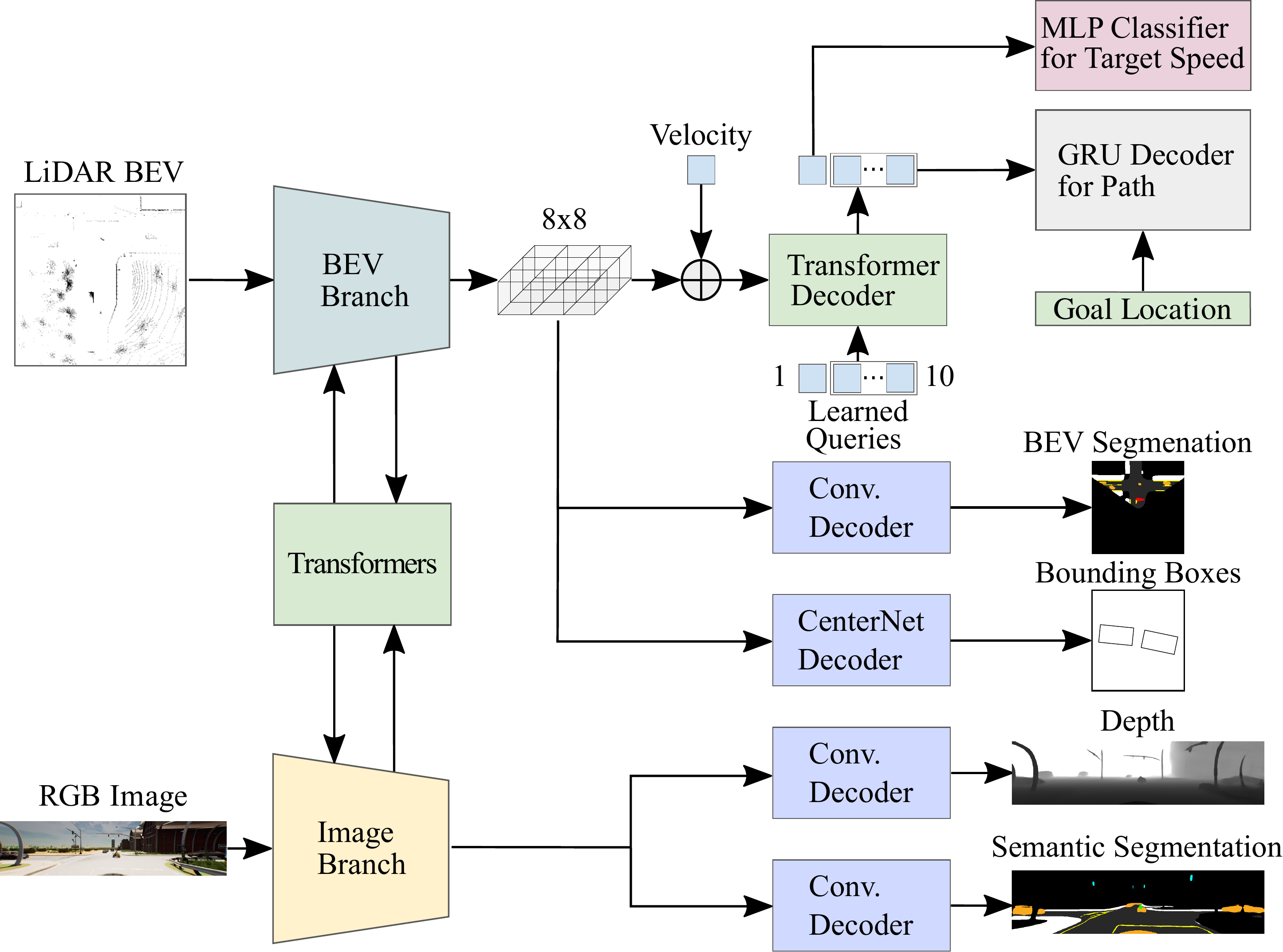}
\end{center}
\vspace{-0.3cm}
\caption{\textbf{TransFuser++ architecture.}}
\label{fig:architecture}
\vspace{-0.1cm}
\end{figure}

\subsection{Scaling up to TransFuser++}

By incorporating these insights, we obtain a significantly improved version of TransFuser~\cite{Chitta2022PAMI} that we call TransFuser++ (\figref{fig:architecture}). We now describe two important implementation changes compared to~\cite{Chitta2022PAMI} that involve scaling.

\begin{table}[ht]
\small
\centering
    \begin{tabular}{c | c | c c | c }
        \toprule
        \textbf{Two stage} & \textbf{Frozen} & \textbf{DS} $\uparrow$ & \textbf{RC} $\uparrow$ & \textbf{Veh} $\downarrow$  \\
        \midrule
        {-} & {-} & {50} \pmsd {3} & {88} \pmsd {1} & {0.83}\\
        $\checkmark$ & {-} & \textbf{54} \pmsd {1} & {92} \pmsd {5} & \textbf{0.79}\\
        $\checkmark$ & $\checkmark$ &  {44} \pmsd {6} & \textbf{97} \pmsd {1} & {1.21}\\
        \bottomrule
    \end{tabular}
    \caption{\textbf{Two stage training.}}
    \label{tab:two_stage}
    \vspace{-0.5cm}
\end{table}

\boldparagraph{Training schedule} We observe no benefits from simply doubling the number of training epochs for a single training stage. However, in \tabref{tab:two_stage}, we double the training time with a two-stage approach. First, we pre-train the encoder with only the perception losses (2D depth, 2D and BEV semantic segmentation, and vehicle bounding boxes, as shown in \figref{fig:architecture}) for the usual number of epochs. We then fine-tune the resulting checkpoint with all losses, after including the GRU decoder and MLP classifier. Initializing the backbone with features that are pre-trained on the auxiliary tasks leads to a 4 DS improvement, consistent with the claims of LAV~\cite{Chen2022CVPRa}. We also experiment with freezing the pre-trained backbone and only training the transformer decoder and its heads in the second stage. This leads to a drop of 10 DS, indicating that end-to-end optimization is important.

\boldparagraph{Dataset scale} Recent work on CARLA~\cite{Renz2022CORL} shows large improvements by scaling the dataset size. However, \cite{Renz2022CORL} considers planning models with privileged inputs on training towns. We investigate the impact of scaling when evaluating on held-out validation towns for end-to-end models. We start with 185k  training samples (as in~\cite{Chitta2022PAMI}) and scale it up by re-running the training routes 3 times with different traffic (555k frames). \tabref{tab:scaling} shows a clear improvement of 6 DS via scaling. This shows that current IL approaches can still benefit from larger datasets. We train for the same amount of epochs, which implies that this improvement comes at the cost of 3$\times$ longer training.

\begin{table}[h]
\small
\centering
    \begin{tabular}{c | c c | c}
        \toprule
        \textbf{Dataset size} & \textbf{DS} $\uparrow$ & \textbf{RC} $\uparrow$ & \textbf{Veh} $\downarrow$ \\
        \midrule
        185k  & {54} \pmsd {1} & {92} \pmsd {5} & {0.79}\\
        555k &  \textbf{60} \pmsd {6} & \textbf{98} \pmsd {1} & \textbf{0.73} \\
        \bottomrule
    \end{tabular}
    \caption{\textbf{Effects of scale.}}
    \label{tab:scaling}
    \vspace{-0.4cm}
\end{table}
\section{Comparison to State of the Art}
\label{sec:results}

\boldparagraph{Methods} This section demonstrates the advantages of TF++ over several state-of-the-art models, which we list below, starting with the most recent.
(1) \textbf{Interfuser} \cite{Shao2022CORL} processes multiple cameras and a BEV LiDAR by encoding each of them individually with a CNN. The resulting feature grids are fed into a transformer and decoded into perception outputs and the path to follow. The network does not predict longitudinal controls. Instead, it uses the BEV bounding boxes and a simple motion forecast (extrapolating historical dynamics) and linearly optimizes for the target speed using heuristically chosen objectives.
(2) \textbf{Perception PlanT} \cite{Renz2022CORL} is a transformer based method trained in two independent stages that uses BEV bounding boxes as an intermediate representation. Its perception is based on TransFuser, and its planner outputs waypoints.
(3) \textbf{TCP} \cite{Wu2022NeurIPS} is a camera-only model with two output representations: waypoints and controls. During test time, it ensembles the two outputs together with a weighted average that changes based on whether the vehicle is turning.
(4) \textbf{TransFuser} \cite{Chitta2022PAMI} fuses perspective cameras with a BEV LiDAR by processing them with individual CNNs and exchanging features using transformers. It uses global average pooling and waypoint outputs, as described in \secref{sec:secrets}.
(5) \textbf{LAV} \cite{Chen2022CVPRa} processes cameras and LiDAR point clouds into an intermediate BEV representation. It trains a planner on this representation that predicts waypoints for the ego vehicle with a NC conditioned GRU followed by a TP conditioned refinement GRU. During training, the planner is also tasked to predict the trajectory of other surrounding vehicles to increase the number of labels. The authors release two versions of this method, which we call LAV v1 and LAV v2.
(6) \textbf{WOR} \cite{Chen2021ICCVa} is an IL method for which the labels are enriched using a reward function to provide dense supervision for all possible actions. Unlike the other baselines, it is conditioned with the NC and does not utilize the TP. WOR is the best NC conditioned baseline that is publicly available.

\begin{table*}[ht]
\small
\centering
    \begin{tabular}{l| c c c | c c c c c c c}
        \toprule
        \textbf{Method} & \textbf{DS} $\uparrow$ & \textbf{RC} $\uparrow$ & \textbf{IS} $\uparrow$ & \textbf{Ped} $\downarrow$ & \textbf{Veh} $\downarrow$ & \textbf{Stat} $\downarrow$ & \textbf{Red} $\downarrow$ & \textbf{Dev} $\downarrow$ & \textbf{TO} $\downarrow$ & \textbf{Block} $\downarrow$ \\
        \midrule
        WOR \cite{Chen2021ICCVa} & {21} \pmsd {3} & {48} \pmsd {4} & {0.56} \pmsd {0.03} & {0.18} & {1.05} & {0.37} & {1.28} & {0.88} & {0.08} & {0.20} \\
        LAV v1 \cite{Chen2022CVPRa} & {33} \pmsd {1} & {70} \pmsd {3} & {0.51} \pmsd {0.02} & {0.16} & {0.83} & {0.15} & {0.96} & {0.06} & {0.12} & {0.45} \\
        Interfuser \cite{Shao2022CORL} & {47} \pmsd {6} & {74} \pmsd {1} & {0.63} \pmsd {0.07} & {0.06} & {1.14} & {0.11} & {0.24} & \textbf{0.00} & {0.52} & \textbf{0.06}\\
        TransFuser \cite{Chitta2022PAMI} & {47} \pmsd {6} & {93} \pmsd {1} & {0.50} \pmsd {0.06} & {0.03} & {2.45} & {0.07} & {0.16} & \textbf{0.00} & \textbf{0.06} & {0.10}\\
        TCP \cite{Wu2022NeurIPS} & {48} \pmsd {3} & {72} \pmsd {3} & {0.65} \pmsd {0.04} & {0.04} & {1.08} & {0.23} & {0.14} & {0.02} & {0.18} & {0.35}\\
        LAV v2 \cite{Chen2022CVPRa} & {58} \pmsd {1} & {83} \pmsd {1} & {0.68} \pmsd {0.02} & \textbf{0.00} & \textbf{0.69} & {0.15} & {0.23} & {0.08} & {0.32} & {0.11}\\
        Perception PlanT \cite{Renz2022CORL} & {58} \pmsd {5} & {88} \pmsd {1} & {0.65} \pmsd {0.06} & {0.07} & {0.97} & {0.11} & {0.09} & \textbf{0.00} & {0.13} & {0.13}\\
        TF++ (ours) & \textbf{69} \pmsd {0} & \textbf{94} \pmsd {2} & \textbf{0.72} \pmsd {0.01} & \textbf{0.00} & {0.83} & \textbf{0.01} & \textbf{0.05} & \textbf{0.00} & {0.07} & \textbf{0.06}\\
        \midrule
        \textit{Expert} & \textit{81} \pmsd {3} & \textit{90} \pmsd {1} & \textit{0.91} \pmsd {0.04} & \textit{0.01} & \textit{0.21} & \textit{0.00} & \textit{0.01} & \textit{0.00} & \textit{0.07} & \textit{0.09} \\
        \bottomrule
    \end{tabular}
    \caption{\textbf{Performance on training towns (Longest6).} Released models, std over 3 evaluations.}
    \label{tab:longest6}
    \vspace{-0.0cm}
\end{table*}

\begin{table*}[ht]
\small
\centering
    \begin{tabular}{l| c c c | c c c c c c c c}
        \toprule
        \textbf{Method} & \textbf{DS} $\uparrow$ & \textbf{RC} $\uparrow$ & \textbf{IS} $\uparrow$ & \textbf{Ped} $\downarrow$ & \textbf{Veh} $\downarrow$ & \textbf{Stat} $\downarrow$ & \textbf{Red} $\downarrow$ & \textbf{Stop} $\downarrow$ & \textbf{Dev} $\downarrow$ & \textbf{TO} $\downarrow$ & \textbf{Block} $\downarrow$ \\
        \midrule
        TransFuser (ours) & {39} \pmsd {9} & {84} \pmsd {7} & {0.46} \pmsd {0.06} & \textbf{0.00} & {0.74} & {1.04} & {0.20} & {1.07} & {0.00} & {0.23} & {0.21}\\
        TCP~\cite{Wu2022NeurIPS} & {58} \pmsd {5} & {85} \pmsd {3} & {0.67} \pmsd {0.06} & \textbf{0.00} & \textbf{0.35} & {0.16} & \textbf{0.01} & {1.05} & {0.00} & {0.19} & {0.19}\\
        TF++ (ours) & \textbf{70} \pmsd {6} & \textbf{99} \pmsd {0} & \textbf{0.70} \pmsd {0.06} & {0.01} & {0.63} & \textbf{0.01} & {0.04} & \textbf{0.26} & \textbf{0.00} & \textbf{0.05} & \textbf{0.00}\\
        \midrule
        \textit{Expert} & \textit{94} & \textit{95} & \textit{0.99} & \textit{0.00} & \textit{0.02} & \textit{0.00} & \textit{0.02} & \textit{0.00} & \textit{0.00} & \textit{0.00} & \textit{0.08}\\
        \bottomrule
    \end{tabular}
    \caption{\textbf{Performance on validation towns (LAV).} Reproduced models, std over 3 trainings (for 3 evaluations each).}
    \label{tab:lav}
    \vspace{-0.5cm}
\end{table*}

\boldparagraph{Benchmarks} We use two benchmarks to evaluate on seen and unseen towns, with scenarios taken from~\cite{Chitta2022PAMI} (type 1,3,4,7,8,9 and 10). \tabref{tab:longest6} compares the performance of systems on the Longest6 benchmark~\cite{Chitta2022PAMI} which consists of 36 routes in training towns 01 to 06. On Longest6, models trained on any data can be evaluated, so we compare against the author-provided models or directly report numbers from the respective papers. The mean and std of three evaluations is reported. For TF++ we train 3 models and report the average result. Longest6 has the advantage that it evaluates in dense traffic and has diverse towns and routes. The drawbacks are that it does not penalize stop sign infractions or measure generalization to new towns. Therefore, we additionally compare performance on validation towns using the LAV \cite{Chen2022CVPRa} routes in \tabref{tab:lav}. These are 4 routes with 4 weathers (16 combined) in Town 02 and 05, which are withheld during training. We re-train methods with the dataset from the corresponding paper 3 times and evaluate each seed 3 times. Reported results are the mean of all runs and the std between the training seeds. We compare against the reported SotA TCP and our reproduced TransFuser baseline on this benchmark and provide additional baselines in the supplementary.

\boldparagraph{Dataset} For TF++ we collect data on the same training routes as in~\cite{Chitta2022PAMI} with an improved expert labeling algorithm (described in the supplementary material). We repeat this 3 times on the same routes with different traffic, as in~\cite{Renz2022CORL}. For validation, we withhold Town 02 and 05 during training, else we train on all towns. In total, we train with 750k frames (550k when withholding Town 02 and 05).

\boldparagraph{Results} We start with the results in training towns (\tabref{tab:longest6}). The best NC conditioned method, WOR, has significantly lower RC than all TP conditioned systems (22 lower than LAV v1). In particular, the route deviations (Dev) are 10 times higher. Current TP conditioned SotA methods achieve (close to) 0 route deviations with driving scores in the range of 47 DS to 58 DS. TF++ outperforms all baselines on Longest6, showing a 19\% relative improvement over the previous SotA Perception PlanT. TF++ achieves close to expert-level performance on all infractions except for vehicle collisions (Veh). On the validation towns (\tabref{tab:lav}) TF++ outperforms our reproduced TransFuser by 31 DS improving all metrics, particularly environmental collisions (Stat). It improves the prior SotA result TCP by 21\%.

\boldparagraph{Runtime} \tabref{tab:runtime} compares the runtime of TF++ and TransFuser which have the same sensor setup. TF++ yields a large DS improvement while being only 14\% slower. 

\begin{table}[ht]
\small
    \centering
    \begin{tabular}{l | c c}
        \toprule
        \textbf{Method} $\uparrow$ & \textbf{DS} $\uparrow$ & \textbf{Time (ms)} $\downarrow$ \\
        \midrule
        TransFuser (ours) & 39 & 44\\
        TF++ (ours) & 70 & 50 \\
        \bottomrule
    \end{tabular}
    \caption{\textbf{Runtime.} We show the runtime per frame in ms averaged over 300 time steps on a single route on a RTX 3090. TF++ outperforms TransFuser by a wide margin despite using a similar compute budget during inference.}
    \label{tab:runtime}
    \vspace{-0.5cm}
\end{table}
\section{Conclusion}

In this work, we show that recent SotA driving models have exceptional route following abilities because they learn a strong bias towards following nearby TPs. Shortcut learning like this is a general phenomenon in deep neural networks~\cite{Geirhos2020NatureMI} and has been observed in the context of autonomous driving for inputs such as velocity~\cite{Codevilla2019ICCV} or temporal frames~\cite{Bansal2019RSS, Wen2020NEURIPS}. We add to this literature by observing shortcut learning with respect to the conditioning signal. While shortcut learning usually has a negative impact on performance, we observe positive (improved recovery) and negative (cutting turns) effects. We show that the negative effects can be mitigated by avoiding global pooling and incorporating data augmentation.

A second commonality in SotA approaches is the use of waypoints as an output representation. We observe that this representation is ambiguous because it predicts a point estimate for multi-modal future velocities. We disambiguate the representation by disentangling future velocities from the deterministic path predictions and classifying target speeds instead. We then weigh target speeds according to their predicted confidence in our controller. Surprisingly, we find that interpolation is helpful for reducing collisions.

We propose TransFuser++ by improving the popular baseline TransFuser with a series of controlled experiments. TF++ is a simple end-to-end method that sets a new state of the art on the LAV and Longest6 benchmarks.

\boldparagraph{Limitations} This study investigates urban driving in CARLA, where all investigated methods drive at relatively low speed ($< 35$km/h). Therefore, problems specific to high-speed driving are not considered. In addition, lanes are free of static obstacles, hence scenarios requiring navigation around them are not included. We point out the strong reliance of current methods on TPs for recovery. This implies a reliance on accurate localization and mapping to obtain these TPs. They are accurately mapped in CARLA, however, this assumption might not hold in real environments.

\boldparagraph{Broader Impact} We aim to make progress towards autonomous driving. This technology, if realized, could have massive societal impact, reducing road accidents, transportation costs and improving the mobility of elderly people. Potential negative implications include a reduction in jobs for human drivers and possible military applications.

\vspace{0.2cm}
\boldparagraph{Acknowledgements} Andreas Geiger and Bernhard Jaeger were supported by the ERC Starting Grant LEGO-3D (850533), the BMWi in the project KI Delta Learning (project number 19A19013O) and the DFG EXC number 2064/1 - project number 390727645. Kashyap Chitta was supported by the German Federal Ministry of Education and Research (BMBF): Tübingen AI Center, FKZ: 01IS18039A. We thank the International Max Planck Research School for Intelligent Systems (IMPRS-IS) for supporting Bernhard Jaeger and Kashyap Chitta as well as Niklas Hanselmann and Katrin Renz for proofreading.

{\small
\bibliographystyle{ieee_fullname}
\bibliography{bibliography_long,bibliography,bibliography_custom}
}
\newpage
\begin{appendices}

\section{Changes to TransFuser}
\label{sec:implementation_details}
Our reproduced TransFuser largely follows \cite{Chitta2022PAMI} but has some minor differences in implementation details described in this section.

\begin{figure*}[t]
\centering
    \begin{subfigure}[b]{0.33\textwidth}
        \centering
        \includegraphics[width=\textwidth]{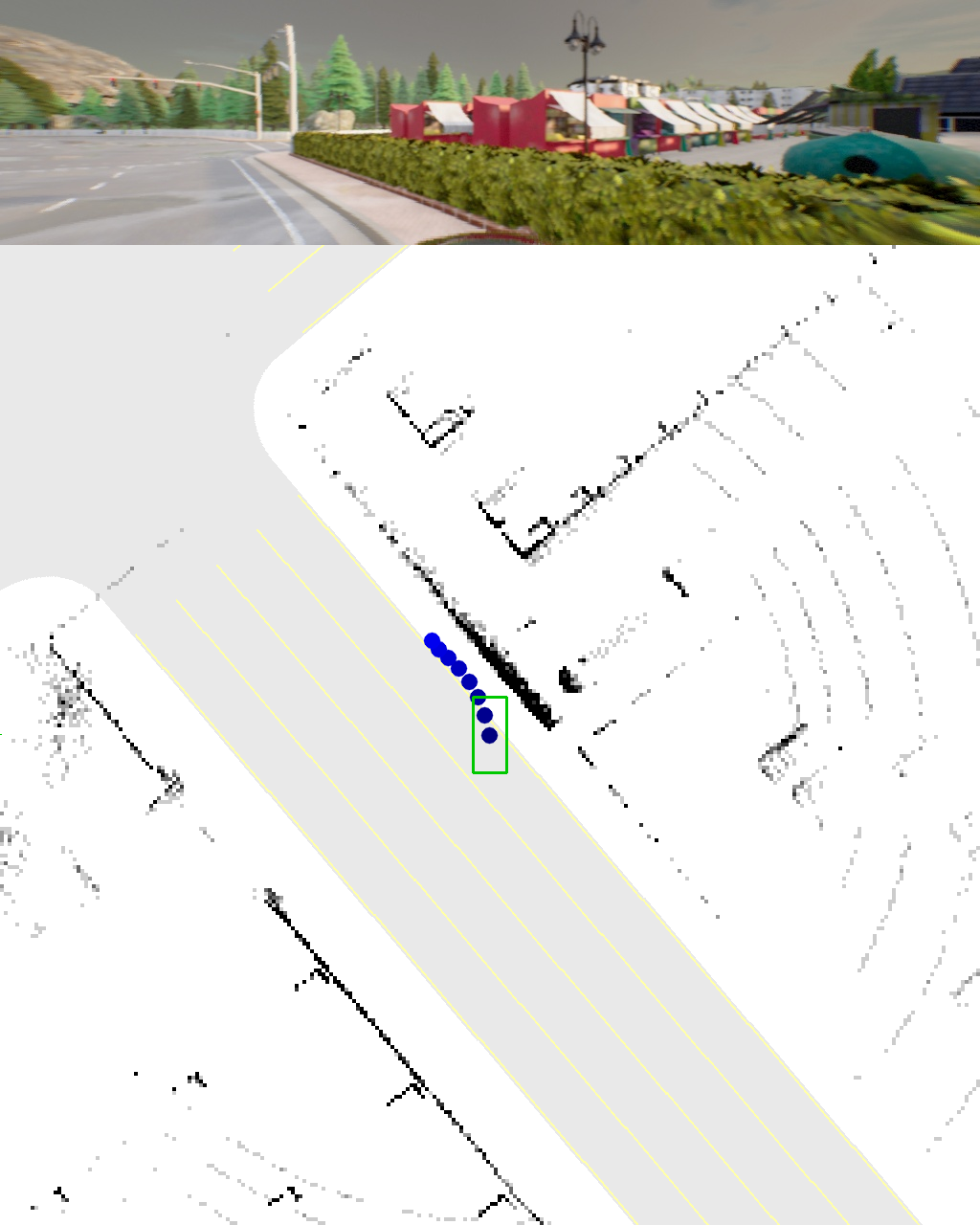}
        \caption{TransFuser (NC conditioned)}
        \label{fig:discrete_target_point_failure_2}
    \end{subfigure}
    \begin{subfigure}[b]{0.33\textwidth}  
        \centering 
        \includegraphics[width=\textwidth]{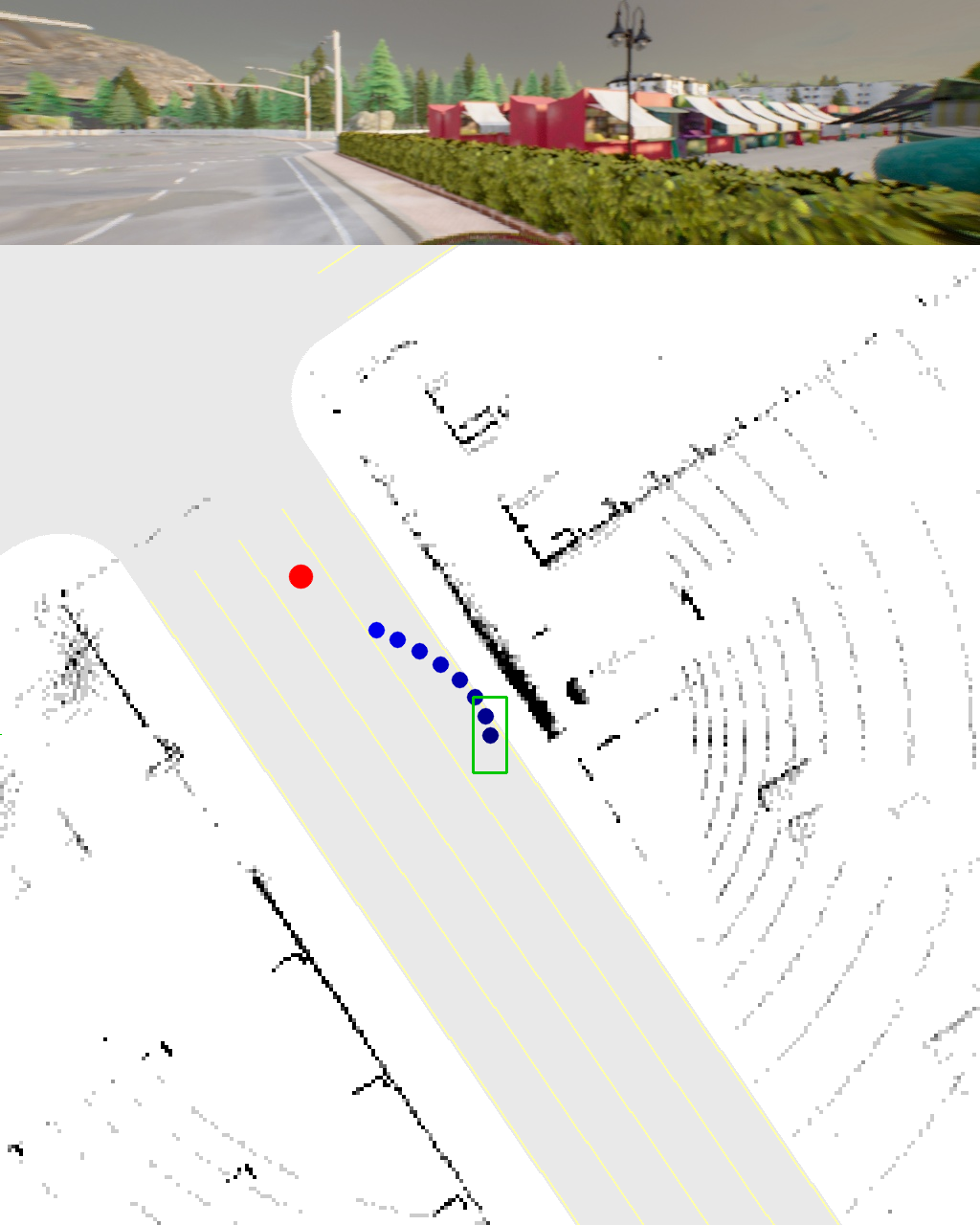}
        \caption{TransFuser (TP conditioned)}
        \label{fig:target_point_success_2_supp}
    \end{subfigure}
    \begin{subfigure}[b]{0.33\textwidth}
        \centering
        \includegraphics[width=\textwidth]{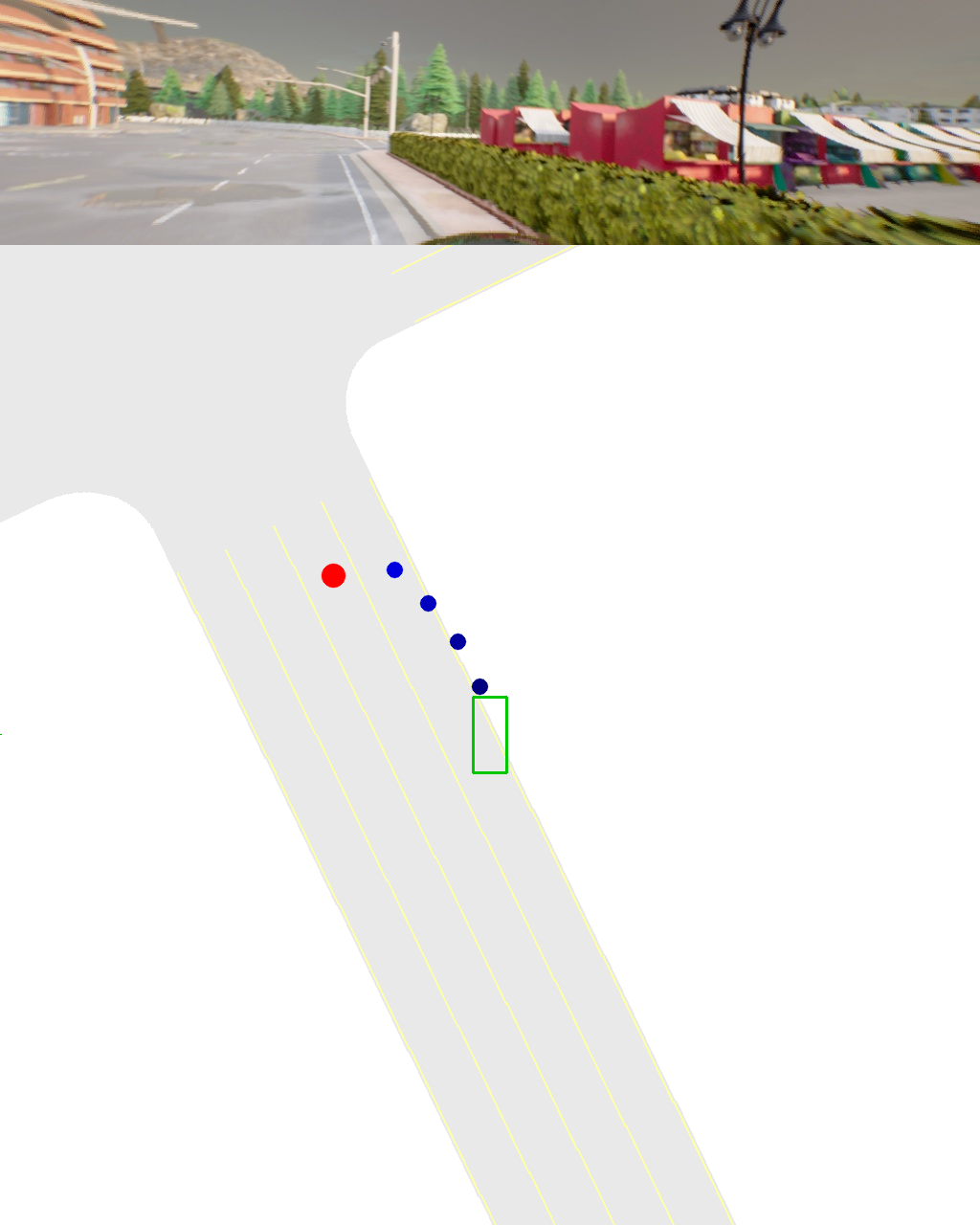}
        \caption{TCP~\cite{Wu2022NeurIPS} (TP conditioned)}
        \label{fig:target_point_success_TCP_2}
    \end{subfigure}
    \vspace{-0.7cm}
   \caption{\textbf{Extrapolation to target point.} Additional example of an unknown situation where TP conditioned methods extrapolate their \textcolor{blue}{waypoints} towards \textcolor{red}{target points}. The TP conditioned models succeed to drive back to the lane, while the discrete conditioned model gets stuck on the sidewalk.}
\label{fig:TP_Recovery_OOD_2}
\end{figure*}

\subsection{Expert}
\label{sec:expert}
For data collection, we follow the common practice of using an automatic labeling algorithm (expert driver) to generate the imitation labels. The auxiliary perception labels are provided directly by the CARLA simulator except for the bird's eye view (BEV) segmentation, for which we follow \cite{Zhang2021ICCV} and render the relevant objects into an HD map. To generate the imitation labels, the expert driver needs to solve planning and control, but can bypass perception using privileged access to the simulator. We build our expert upon the model predictive control (MPC) approach of \cite{Chitta2022PAMI}. Lateral control is done by following the next point (at least 3.5 meters away) in a path, created by an A$^\star$ algorithm, with a PID controller. Longitudinal control is done via MPC that differentiates between 4 target speeds. For regular driving, we use 8 m/s (double the target speed compared to \cite{Chitta2022PAMI}). Inside intersections, we slow down to 5 m/s. 

\boldparagraph{Collision avoidance} The target speed is set to 0 m/s when the MPC algorithm predicts a collision. We predict collisions similarly to \cite{Chitta2022PAMI}. They approximate all agents' future positions by iteratively unrolling a kinematic bicycle model \cite{Chen2021ICCVa}. Actions for other cars are set to be the same as the current time step while the ego agent's own action is approximated by using a PID controller to follow the A$^\star$ path. In case the ego agent's bounding box overlaps with another agent's bounding box at future time step $t$, a collision is predicted. To keep a safety distance to the leading vehicle, the MPC expert in \cite{Chen2021ICCVa} additionally has a static bounding box in front of the ego agent that predicts a collision if it intersects with any other agent. Because our agent has double the maximum driving speed, we need to keep a larger safety distance. Using a static area would not suffice here, as the static bounding box will become so large that it sticks into the opposing lane during turns, causing unnecessary braking. Instead, we approximate the area where the expert would end up if it performed a full brake (after 1 meter of driving) at his current speed $s$. The distance to stop $d$ is approximated with:
\begin{equation}
    d = 0.5 * (\frac{s * 3.6}{10.0})^2 + 2.5
\end{equation}
We again use the kinematic bicycle model to compute where the expert will be after this distance. If the bounding box of the expert at that time step intersects with any bounding boxes for the current time step, we set the target speed to 0. This approach has the advantage that the safety area follows the road and increases with increasing agent speed.
A second problem coming from the higher driving speed is scenario 3 where a pedestrian runs in front of the vehicle. The expert will stop as soon as the pedestrian attempts to move forward, but when driving at 8 m/s this is already too late to prevent the collision. To solve this problem, we preemptively slow down to 2 m/s whenever a pedestrian is within a 30-meter radius in front of the vehicle. 

\boldparagraph{Traffic rules} Stop signs infractions are addressed by slowing down to 2 m/s when the safety area intersects with the stop sign trigger area (provided by the simulator) and stopping once the car is on the stop sign trigger area. Red lights are resolved by setting the target speed to 0 when the safety area or vehicle intersects with the entrance of an intersection and the corresponding traffic light is active.
\begin{table}[ht]
\small
\centering
    \begin{tabular}{l | c c | c c}
        \toprule
        \textbf{Method} & \textbf{DS} $\uparrow$ & \textbf{RC} $\uparrow$ & \textbf{Veh} $\downarrow$ & \textbf{Block} $\downarrow$\\
        \midrule
        MPC Expert \cite{Chitta2022PAMI} & {77} \pmsd {2} & {89} \pmsd {1} & {0.28} & {0.13} \\ 
        MPC Expert (ours) & \textbf{81} \pmsd {3} & \textbf{90} \pmsd {1} & \textbf{0.21} & \textbf{0.09}\\ 
        \bottomrule
    \end{tabular}
    \caption{\textbf{Expert comparison on Longest6}}
    \label{tab:expert_comparison}

\end{table}

\boldparagraph{Performance} We show in \tabref{tab:expert_comparison} that our expert outperforms the baseline by 4 DS. This represents a higher upper bound for our imitation learning models. However, since no method can currently reach that upper bound, we see no direct improvement from the better expert (see \tabref{tab:transfuser_reproduced}).

\subsection{Dataset}
We generate our training dataset by executing the expert described in \secref{sec:expert} on the training routes from \cite{Chitta2022PAMI} and storing every frame (20 FPS). Weathers are randomized per route instead of per frame to avoid exposure problems. During training, we train on every fifth frame, leading to an effective FPS of 4. For the scaling experiment, we rerun data collection on each route 3 times. This randomizes weathers and traffic (the traffic manager in CARLA 0.9.10 is not deterministic) but the environment is the same.

For the final model in the main paper we recollect the data at 4 FPS and train on every frame. The model sees equivalent data, but this makes the second dataset is 5x smaller and hence easier to release. To reduce storage requirements further, we compress the camera images with JPG (we add the compression during inference as well to avoid distribution shifts), perception labels with PNG (depth maps are stored at 8 bit resolution), text files with zip, and LiDAR point clouds with LASzip \cite{Isenburg2013PERS} (which compresses point clouds much better than standard zip). We store full 360\textdegree\ LiDAR sweeps and realign all points into the coordinate system of the current frame. Since the expert is not a perfect driver, we log the driving score of each training route and only train on routes with 100 DS. The labels for the disentangled path are generated by storing the next 10 2D points (spaced 1 meter apart) from the A$^\star$ path the expert is following.

\begin{figure*}[t]
\centering
    \begin{subfigure}[b]{0.33\textwidth}
        \centering
        \includegraphics[width=\textwidth]{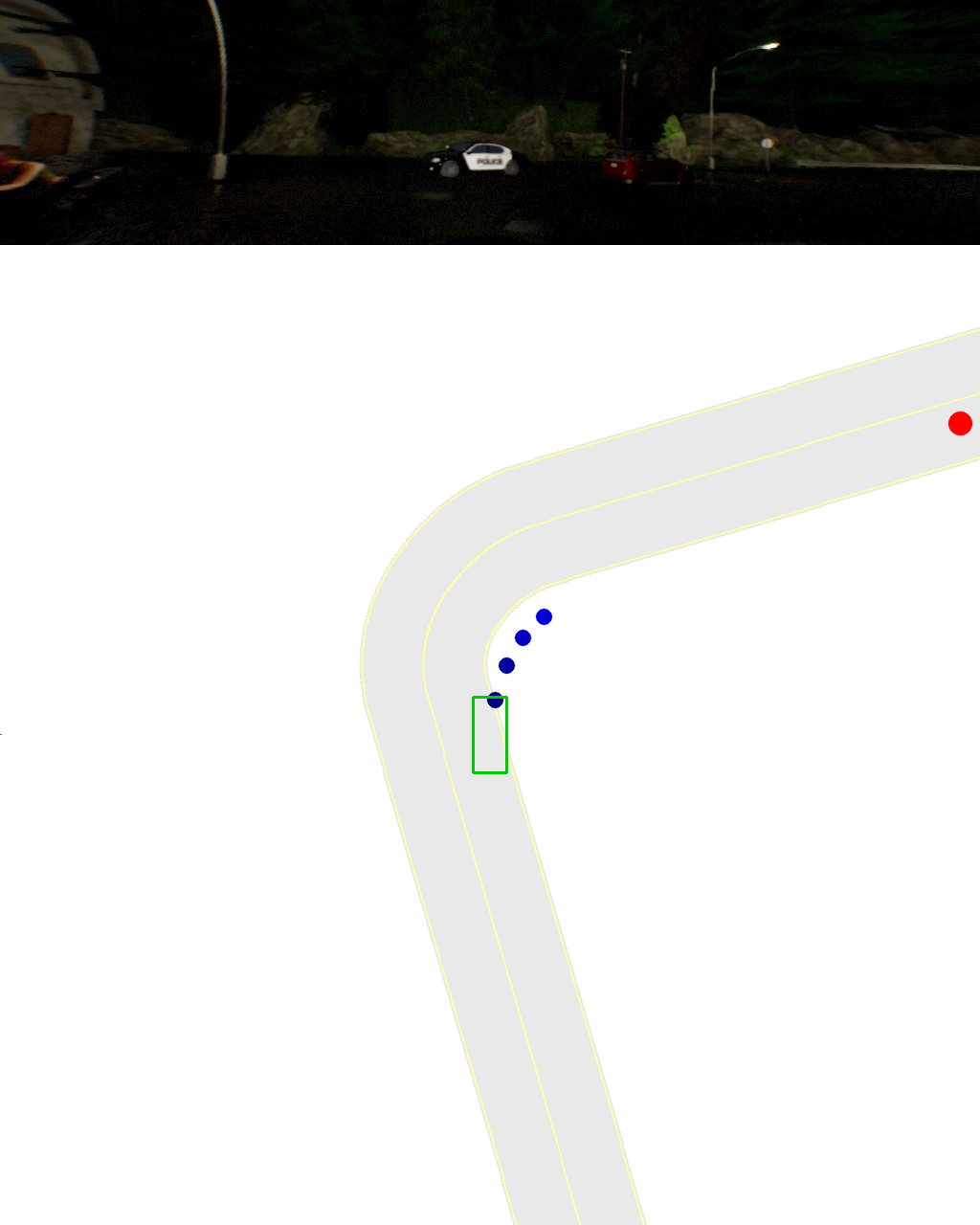}
        \caption{TCP~\cite{Wu2022NeurIPS} follows a shortcut.}
        \label{fig:target_point_failure_tcp_2}
    \end{subfigure}
    \begin{subfigure}[b]{0.33\textwidth}
        \centering
        \includegraphics[width=\textwidth]{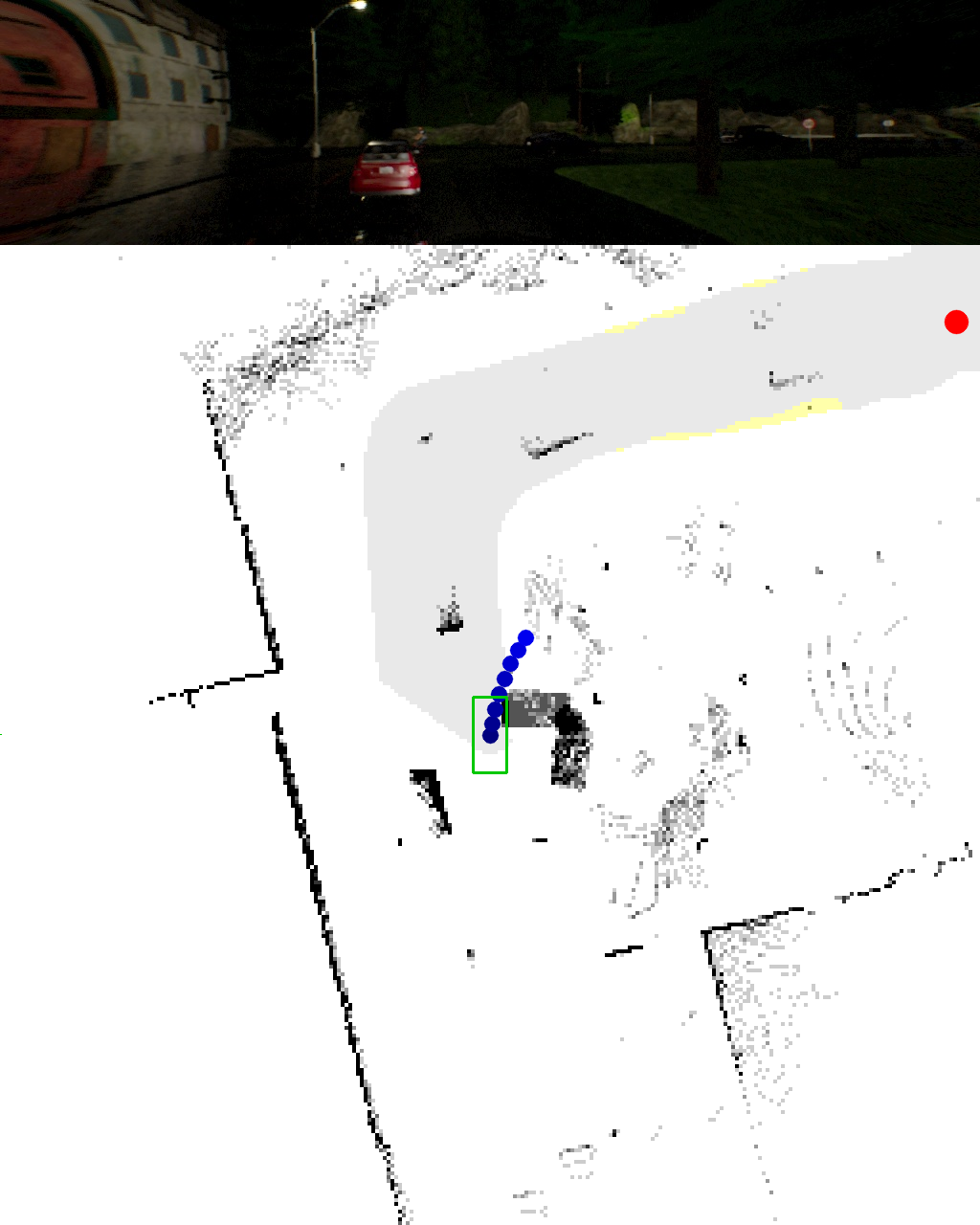}
        \caption{TransFuser follows a shortcut.}
        \label{fig:target_point_failure_2}
    \end{subfigure}
    \begin{subfigure}[b]{0.33\textwidth}  
        \centering 
        \includegraphics[width=\textwidth]{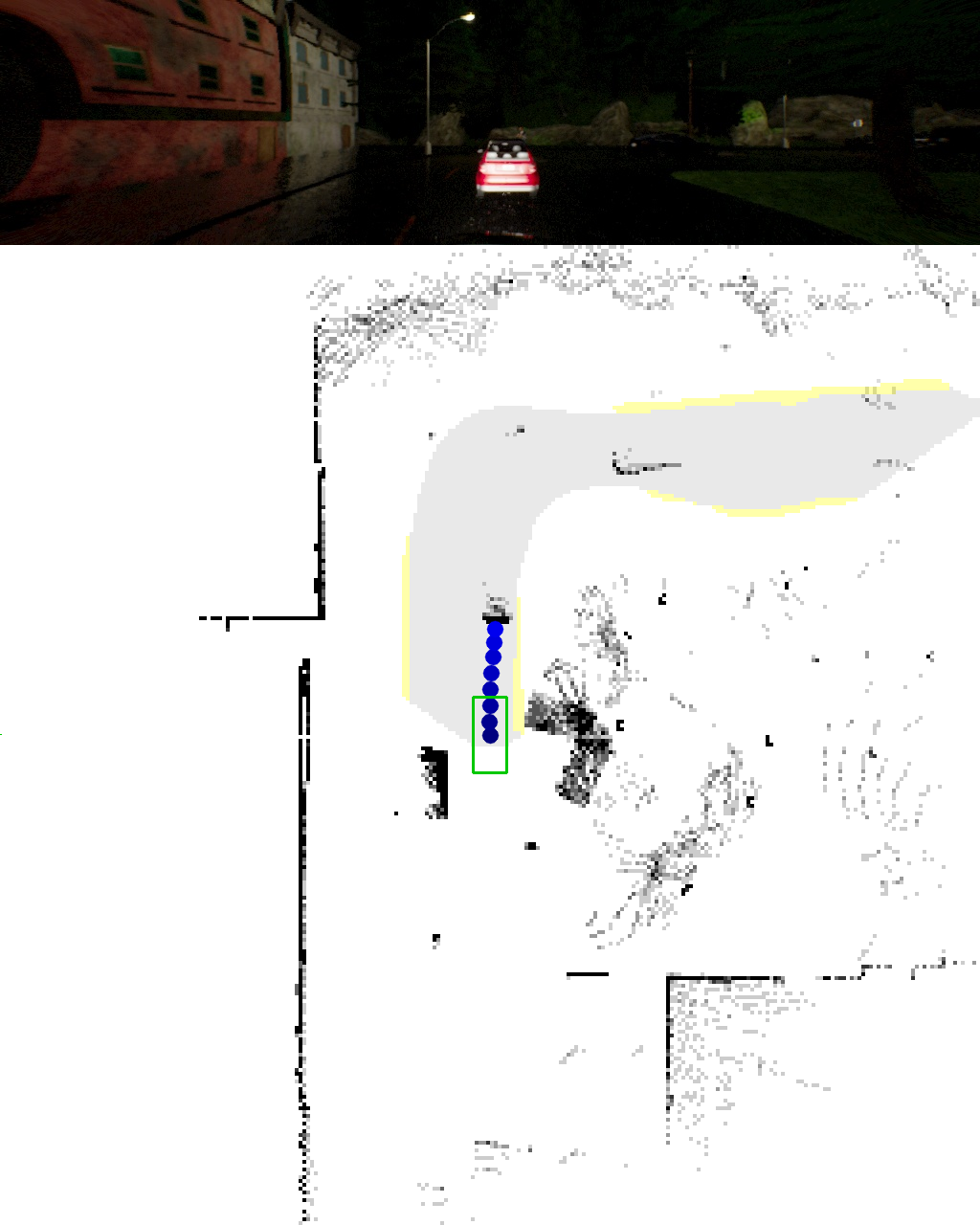}
        \caption{A transformer decoder fixes this.}
        \label{fig:tf_decoder_success_2}
    \end{subfigure}
    \vspace{-0.7cm}
   \caption{\textbf{Target point shortcut.} When TP conditioned methods extrapolate to spatially distant waypoints, they incur large steering errors. Replacing global average pooling in TransFuser with a cross-attention mechanism mitigates the issue.}
\label{fig:target_point_2}
\end{figure*}

\subsection{Training and Architecture}
\label{sec:changes_to_tf}
\boldparagraph{Training} We use the same loss functions as in \cite{Chitta2022PAMI} for all 5 outputs. For the new target speed classification task we use a class frequency weighted cross-entropy loss with label smoothing 0.1. As classes we use the 4 different target speeds of the expert (see \secref{sec:expert}). We observe some failure cases where the target speed predictions are overconfident, and the car starts braking a few frames too late when stopping behind another vehicle, leading to close rear collisions. Label smoothing addresses this problem because it makes the car drive a bit slower (due to the probability weighting). This gives the car a few more frames time to brake and avoids these collisions. To make this more explicit, we do not use label smoothing in the final model in the main paper, and instead reduce the target speeds by 2 m/s during inference, achieving the same effect.

During training, each individual loss is added together with the same weight. The loss weights are normalized to sum to 1. We use AdamW \cite{Loshchilov2019ICLR} with amsgrad \cite{Reddi2018ICLR} and a slightly higher learning rate of 0.0003 which is reduced by a factor of 10 after epoch 30. We train for 31 epochs and always use the last epoch as our model. All models in Section 3 of the main paper are trained on 8 2080ti GPUs with distributed data parallel and total batch size of 48. For the dense dataset, we subsample by a factor of 5 but shift the first frame by the GPU index, so that all the frames are seen during training (a subtle form of data augmentation, close by frames are mostly redundant).

The final model is trained with 4 A100 GPUs and batch size 128 to speed up training. We also add standard color augmentations to the camera and a discrete conditioning concatenated with the velocity input, so that the target speed branch also has a conditioning signal. This is conceptually more sound, but we observed no significant impact on driving performance. Our path labels have a subtle ambiguity due to conversion from a path in global coordinates to the local ego coordinate system (points can be shifted by 1 meter depending where the car is on the path). To resolve this, we linearly interpolate between the stored points at 1 meter distances in ego coordinates during training. This makes the predictions look qualitatively more consistent, but again has no significant impact on driving score.

\boldparagraph{Architecture} For the BEV semantic segmentation we predict 11 classes (unlabeled, road, sidewalk, non cross-able lane marker, cross-able lane marker, stop signs, traffic light stop line green / yellow / red, vehicle, pedestrians) \cite{Zhang2021ICCV}. We only predict pixels that are visible in the camera, since some classes need RGB features. For the perspective segmentation we predict 7 classes (unlabeled, road, sidewalk, lane markings, traffic light, vehicle, pedestrian). For the bounding boxes we predict 4 classes (red and yellow traffic light stop line, stop sign, vehicles and pedestrians). Traffic lights and stop sign boxes are only predicted if they affect the ego vehicle. Vehicles and pedestrians only when they are in the LiDAR range. Architecture wise we add 1x1 convolution layers in the LiDAR branch before each transformer that match the channel dimension of the LiDAR to the one in the image branch (allows using different backbones for the two branches, though we keep using RegNetY-3.2GF \cite{Radosavovic2020CVPR} for both). Our reproduced TransFuser predicts 8 waypoints, each placed 250 ms apart (up to 2 seconds into the future).

\boldparagraph{Sensors} We follow \cite{Wu2022NeurIPS} and use a single high resolution camera. It has a 110\textdegree\ horizontal field of view (FOV), is mounted at (-1.5, 0.0, 2.0) and has a resolution of 256 x 1024. Our LiDAR is mounted at (0.0,0.0, 2.5) and has a full 360\textdegree\ FOV. LiDAR points are realigned into the current vehicle coordinate system by using the filter described in \secref{sec:localization}. The LiDAR points are voxelized into a 256x256 grid representing a 64 x 64 meter area with the vehicle at its center. Each pixel covers a 0.25 x 0.25 meter area. We remove the LiDAR ground plane by removing all points with a height of less than 0.2 meters. 

\boldparagraph{Performance} The reproduced TransFuser has a slightly lower DS to the original on the Longest6 benchmark as shown in \tabref{tab:transfuser_reproduced}. The difference in driving score likely is a result of the driving score weighting the dissimilar failure modes differently. Note, that our reproduced TransFuser is a single model, whereas the original result was from an ensemble of 3 models. 
\begin{table}[ht]
\small
\centering
    \begin{tabular}{l | c c | c c}
        \toprule
        \textbf{Method} & \textbf{DS} $\uparrow$ & \textbf{RC} $\uparrow$ & \textbf{Veh} $\downarrow$ & \textbf{Stat} $\downarrow$ \\
        \midrule
        TransFuser (ours) & {40} & {82} & \textbf{1.17} & {0.57} \\ 
        TransFuser \cite{Chitta2022PAMI} & \textbf{47} & \textbf{93} & {2.45} & \textbf{0.07} \\ 
        \bottomrule
    \end{tabular}
    \caption{\textbf{Reproduced TF vs original on Longest6}}
    \label{tab:transfuser_reproduced}
\end{table}

\subsection{Localization}
\label{sec:localization}
We localize our vehicle with a GNSS sensor. The GNSS signal has Gaussian noise applied to it, leading to average localization errors of $\sim$0.7 meters. Like prior work \cite{Chitta2022PAMI, Chen2022CVPRa} we use a filtering algorithm to reduce the noise. In particular, we use an Unscented Kalman Filter (UKF) \cite{Merwe2000IEEE, Labbe2020} with Van der Merwe’s scaled sigma point algorithm \cite{Merwe2004Thesis}. As its model of the car, our UKF uses the same kinematic bicycle model as the expert \cite{Chen2021ICCVa}. The filter tracks the position, orientation and speed of the ego vehicle. The parameters of the filter are tuned manually by reducing localization error on a small dataset consisting of ground truth localization paired with GNSS signals (similar to tuning hyperparameters on a validation set in supervised learning). The filter reduces the average localization error to below $\sim$0.1 meters.

\begin{figure*}[t]
\centering
    \begin{subfigure}[b]{0.33\textwidth}
        \centering
        \includegraphics[width=\textwidth]{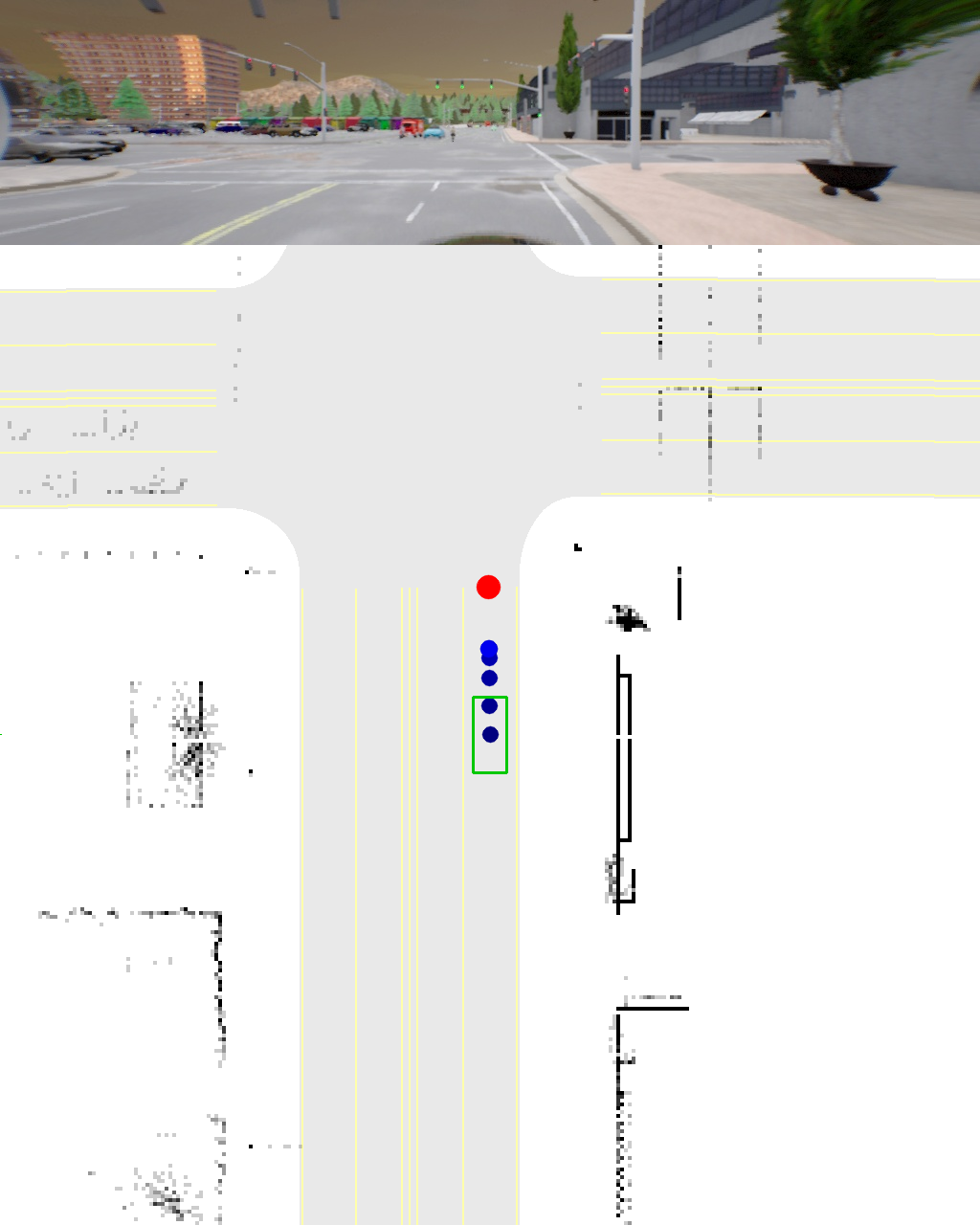}
        \caption{Slowing down at a green light.}
        \label{fig:conservative_mode_2}
    \end{subfigure}
    \begin{subfigure}[b]{0.33\textwidth}  
        \centering 
        \includegraphics[width=\textwidth]{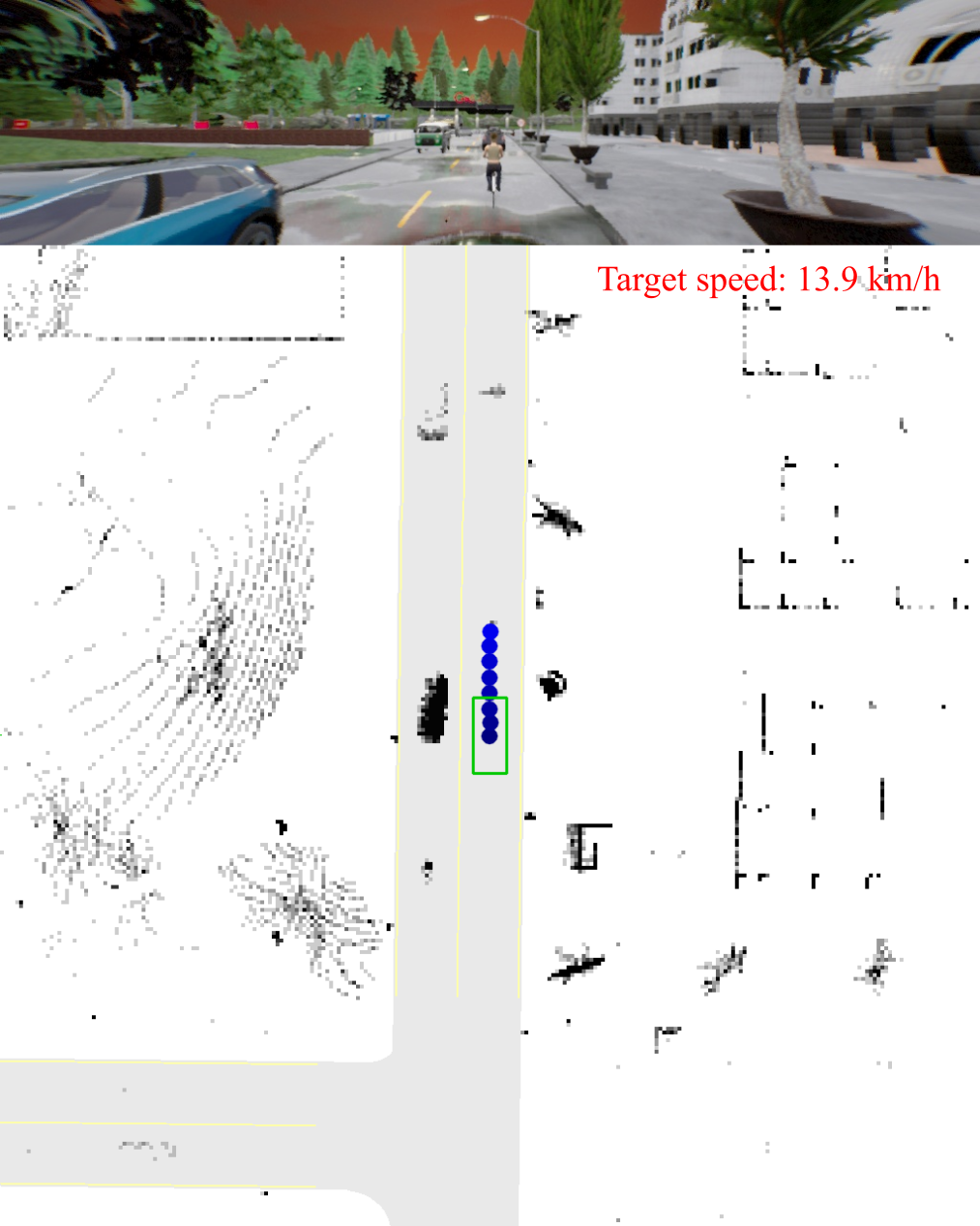}
        \caption{Slowing down behind a cyclist.}
        \label{fig:interpolation_following_wp}
    \end{subfigure}
    \begin{subfigure}[b]{0.33\textwidth}  
        \centering 
        \includegraphics[width=\textwidth]{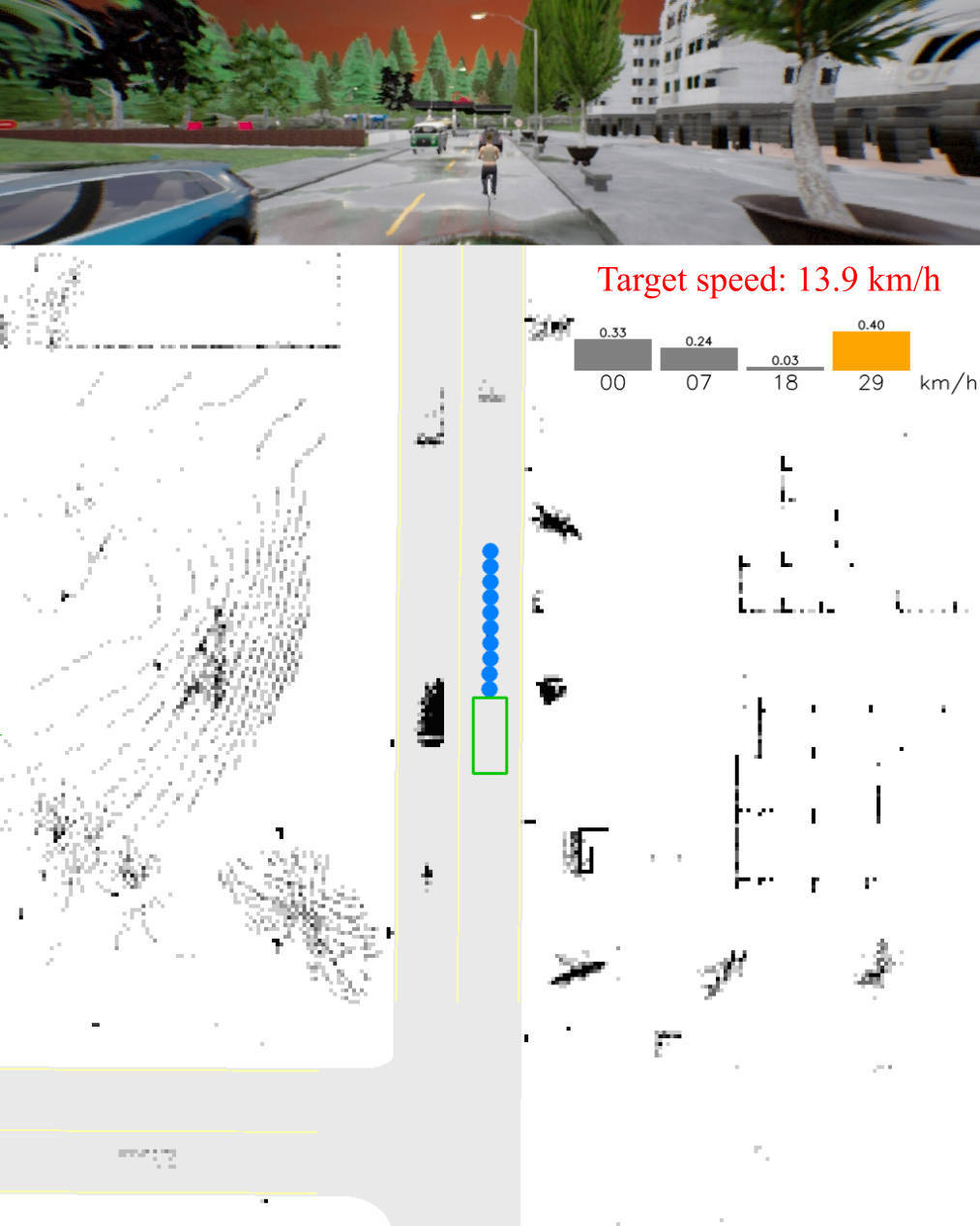}
        \caption{Disentangling route and target speed.}
        \label{fig:interpolation_following_discrete}
    \end{subfigure}
    \vspace{-0.7cm}
   \caption{\textbf{Waypoints are ambiguous.} The model's output representation forces it to predict a single mode for future velocities.}
\label{fig:waypoint_problem_2}
\end{figure*}

\subsection{PID Controller}
In \cite{Chitta2022PAMI} the PID controller converts the waypoints into steering by computing the car's angle towards the average between the first two waypoints. The angle is then input to a PID controller to minimize. Computing the angle based on entangled waypoints makes the steering angle towards the path depend on the predicted speed of the vehicle. This works well at the low driving speeds of \cite{Chitta2022PAMI} but becomes hard to tune when driving at higher (and more diverse) speeds like our expert. Instead, we follow the first waypoint that is at least a certain aim distance $a$ away from the center of the car (or the last one). We keep $a$ similar to \cite{Chitta2022PAMI} and use $a=2.25$ when driving slower than 5.5 m/s (inside intersections) and $a=3.0$ when driving faster. Longitudinal control is kept similar, we use the velocity between the waypoint 0.5 second into the future and 1 second into the future as target speed. The models using the disentangled waypoint representation do not require tuning an additional controller. Since we directly predict the input to the expert's PID controller, we can reuse the same PID controllers from the expert (see \secref{sec:expert}), which we know work well.

The task of yielding to stop signs is defined in CARLA to stop the vehicle on a STOP paining printed on the road. This stop painting is largely occluded by the motor hut once the car is on the sign because our camera is mounted at the back roof of the vehicle. We observe many cases where TF++ detects the stop sign initially and slows down but starts driving again once the painting becomes occluded. We can address this problem in the controller because TF++ detects stop sign bounding boxes in BEV as auxiliary task. For our final model, we keep the last detected stop sign in a buffer and transform it to the current ego coordinate system using the UKF motion estimation. If the car is on the stop sign bounding box we set the action to brake in the controller until the stop sign is cleared.
\begin{table}[h]
\small
\centering
    \begin{tabular}{c | c c | c }
        \toprule
        \textbf{Stop Contr.} & \textbf{DS} $\uparrow$ & \textbf{RC} $\uparrow$ & \textbf{Stop} $\downarrow$ \\
        \midrule
        - & {60} \pmsd {5} & 99 \pmsd {1} & {1.35} \\
        \checkmark & \textbf{70} \pmsd {6} & \textbf{99} \pmsd {0} & \textbf{0.26}\\
        \bottomrule
    \end{tabular}
    \caption{\textbf{Stop sign controller.}}
    \label{tab:stop_controller}
    \vspace{-0.3cm}
\end{table}
\tabref{tab:stop_controller} shows the impact of the stop sign controller change on the validation routes. TF++ detects most stop signs and the controller is able to reduce Stop infractions by 5x, leading to a +10 DS improvement.

\section{TransFuser++ Implementation Details}

\subsection{Attention Pooling Implementation}
The core implementation of our attention pooling with a transformer decoder is similar to \cite{Shao2022CORL}. We use the 8x8 features coming from the BEV branch and reduce the number of channels (1512 with our RegNetY-3.2GF backbone) to 256 with a 1x1 convolution. We add a sinusoidal positional embedding to the features and flatten them afterward. The velocity input, we normalize with a 1D Batchnorm and embed it with a 2 layer MLP to 256 features. Afterward, we add a learnable positional encoding. The velocity token then gets concatenated to the BEV tokens. These tokens are then processed by a standard Transformer decoder \cite{Vaswani2017NIPS} that has 8 heads, gelu activation function \cite{Hendrycks2016ARXIV}, 6 layers and uses layernorm \cite{Ba2016ARXIV}. We use 1 learned queries for every predicted waypoint and 1 additional one if we predict target speeds. The tokens are then fed as inputs into a GRU, whose hidden state is initialized with an (MLP embedded) target point. A linear layer converts the output to 2 dimensions, and a cumulative sum is applied to the final vector (forcing the network to predict offsets from the first point).

\subsection{Data Augmentation}
When collecting data with the expert driver, we mount an additional camera on the vehicle that collects augmented frames and labels at every time step. This means for every frame in the dataset, we have an augmented counterpart. We shift the camera by $\pm 1$ meter to the left or right (of the vehicle) and rotate it by $\pm 5^{\circ}$ around the yaw axis. The particular values are drawn from a uniform distribution. During training, we load the perturbed camera with $50\%$ probability. We transform the 2D way point labels and other data such that they have the perturbed camera as center. The shifted and rotated waypoint labels do not describe an actual recovery trajectory that an expert would take, and are instead the original waypoints in the center of the lane. However, the PID controller will steer the car back to the center of the lane, when the network predicts these augmented waypoints, achieving a similar effect. The range of errors the network is trained to recover from equals the range of the perturbations. To generate the perturbed camera, we need access to a rendering engine (CARLA in our case). On real data, a similar effect can be achieved using novel view synthesis~\cite{Amini2022ICRA}.
\begin{table*}[ht]
\small
\centering
    \begin{tabular}{l| c c c | c c c c c c c c}
        \toprule
        \textbf{Method} & \textbf{DS} $\uparrow$ & \textbf{RC} $\uparrow$ & \textbf{IS} $\uparrow$ & \textbf{Ped} $\downarrow$ & \textbf{Veh} $\downarrow$ & \textbf{Stat} $\downarrow$ & \textbf{Red} $\downarrow$ & \textbf{Stop} $\downarrow$ & \textbf{Dev} $\downarrow$ & \textbf{TO} $\downarrow$ & \textbf{Block} $\downarrow$ \\
        \midrule
        LAV v2 \cite{Chen2022CVPRa} & {27} \pmsd {1} & {98} \pmsd {1} & {0.27} \pmsd {0.01} & \textbf{0.00} & \textbf{0.35} & {0.04} & {2.89} & {1.42} & \textbf{0.00} & {0.09} & \textbf{0.00}\\
        Perception PlanT~\cite{Renz2022CORL} & {37} \pmsd {5} & {86} \pmsd {7} & {0.45} \pmsd {0.09} & \textbf{0.00} & {0.92} & {0.31} & {0.09} & {1.87} & \textbf{0.00} & {0.25} & {0.14}\\
        TransFuser (ours) & {39} \pmsd {9} & {84} \pmsd {7} & {0.46} \pmsd {0.06} & \textbf{0.00} & {0.74} & {1.04} & {0.20} & {1.07} & \textbf{0.00} & {0.23} & {0.21}\\
        TCP~\cite{Wu2022NeurIPS} & {58} \pmsd {5} & {85} \pmsd {3} & {0.67} \pmsd {0.06} & \textbf{0.00} & \textbf{0.35} & {0.16} & \textbf{0.01} & {1.05} & \textbf{0.00} & {0.19} & {0.19}\\
        TF++ (ours) & \textbf{70} \pmsd {6} & \textbf{99} \pmsd {0} & \textbf{0.70} \pmsd {0.06} & {0.01} & {0.63} & \textbf{0.01} & {0.04} & \textbf{0.26} & \textbf{0.00} & \textbf{0.05} & \textbf{0.00}\\
        \midrule
        TF++ (all towns) & {90} \pmsd {1} & {99} \pmsd {1} & {0.91} \pmsd {0.02} & {0.00} & {0.18} & {0.00} & {0.05} & {0.00} & {0.00} & {0.02} & {0.00}\\
        \textit{Expert} & \textit{94} & \textit{95} & \textit{0.99} & \textit{0.00} & \textit{0.02} & \textit{0.00} & \textit{0.02} & \textit{0.00} & \textit{0.00} & \textit{0.00} & \textit{0.08}\\
        \bottomrule
    \end{tabular}
    \caption{\textbf{Performance on validation towns (LAV).} Reproduced models, std over 3 trainings and 3 evaluations.}
    \label{tab:lav_additional}
\end{table*}

\section{Additional Results}
\label{sec:additional_results}

\subsection{Longest6 Ablations}
\label{sec:longest6_ablations}
As a sanity check, we test our additions to TransFuser by repeating the experiments on the training towns (Longest6) while training with all data. The results are presented in \tabref{tab:longest6_ablation}.
\begin{table}[ht]
\small
\centering
    \begin{tabular}{l | c c | c c}
        \toprule
        \textbf{Method} & \textbf{DS} $\uparrow$ & \textbf{RC} $\uparrow$ & \textbf{Veh} $\downarrow$ & \textbf{Stat} $\downarrow$ \\
        \midrule
        TransFuser (ours) & {40} \pmsd {3} & {82} \pmsd {2} & {1.17} & {0.57} \\ 
        + transformer decoder & {57} \pmsd {3} & {90} \pmsd {3} & {0.93} & {0.19} \\ 
        + data augmentation &  {66} \pmsd {4} & {94} \pmsd {2} & {0.64} & {0.07} \\
        + disentangled &  {64} \pmsd {4} & {96} \pmsd {1} & {0.88} & {0.02} \\
        + two stage &  {67} \pmsd {2} & {96} \pmsd {1} & {0.82} & {0.01} \\
        + 3x data &  {69} \pmsd {1} & \textbf{97} \pmsd {1} & {0.79} & \textbf{0.00} \\
        \midrule
        TF++ WP (ours) & {70} \pmsd {4}& {94} \pmsd {2} & {0.66} & {0.01} \\
        + ensemble & \textbf{73} & \textbf{97} & \textbf{0.56} & {0.01}\\
        \bottomrule
    \end{tabular}
    \caption{\textbf{Longest6 Ablations.} Each row has the previous one as baseline. Std over 3 training and 3 evaluation runs.}
    \label{tab:longest6_ablation}
\end{table}
We start with our reproduced version of TransFuser and iteratively add 1 change in every row. All results are the average of 3 training seeds, each evaluated 3 times. The reported standard deviation is across the training seeds. The main problems of the reproduced TransFuser are high collisions with other vehicles (Veh) and collisions with the static environments (Stat). Replacing the global average pooling and MLP with a Transformer decoder improves the driving score by 17 points, improving both vehicle collisions and collisions with the environment (indicating improved steering). Adding shift and rotation augmentations has a similar effect, improving the driving score by +9 and reducing collisions. The disentangled representation has a similar effect than on the validation routes, increasing vehicle collisions and reducing collisions with the environment. Since Longest6 has denser traffic, the increased vehicle collisions have a larger impact here, leading to an overall decrease in 2 DS. Two stage training improves DS by +3 and scaling the dataset by 3x leads to a result of 69 DS. 
Since the waypoint representation is slightly better on longest6, we additionally report the result of a variant called TF++ WP in \tabref{tab:longest6_ablation}. It was trained with the released dataset and uses waypoints as output representation, but is otherwise identical to TF++. TF++ WP again has a slightly higher DS of +1. We also report the result of an ensemble of the 3 training seeds following prior work \cite{Chitta2022PAMI}. The ensemble reduces vehicle collisions by 0.10 and leads to a SotA result of 73 DS on longest6.

\subsection{Brake Threshold}
The confidence weighted PID controller from the models using target speed prediction has a hyperparameter that determines at which confidence threshold the action is set to full brake. In \tabref{tab:brake_threshold} we investigate the effects of different thresholds on validation towns.
\begin{table}[ht]
\small
\centering
    \setlength{\tabcolsep}{3pt}
    \begin{tabular}{c | c c c | c}
        \toprule
        \textbf{Threshold} & \textbf{DS} $\uparrow$ & \textbf{RC} $\uparrow$ & \textbf{IS} $\uparrow$ & \textbf{Veh} $\downarrow$ \\
        \midrule
        50\% &  {60} \pmsd {6} & \textbf{98} \pmsd {1} & {0.61} \pmsd {0.06} & {0.73} \\
        40\% &  {58} \pmsd {1} & {96} \pmsd {5} & {0.61} \pmsd {0.04} & {0.78} \\
        33\% &  \textbf{61} \pmsd {3} & {96} \pmsd {4} & \textbf{0.64} \pmsd {0.07} & \textbf{0.61} \\
        25\% &  {59} \pmsd {3} & {95} \pmsd {6} & {0.63} \pmsd {0.08} & {0.64} \\
        \bottomrule
    \end{tabular}
    \caption{\textbf{Brake threshold.}}
    \label{tab:brake_threshold}
\end{table}

Lower thresholds reduce vehicle collisions at the cost of lower route completion due to false positive braking. Overall, the choice of threshold is robust, only changing by 3 DS between the best and worst tested threshold. We use the default threshold of 50\% for all validation town experiments. On Longest6, the brake threshold has a slightly larger impact because the dense traffic puts a higher focus on collision avoidance. We therefore use the threshold 33\% for Longest6 experiments.

\subsection{Additional Baselines}
We report 3 additional results on the validation towns in \tabref{tab:lav_additional}.
TF++ (all towns) is TF++, but its training includes the validation towns. This is not a fair comparison to other methods, and rather serves to illustrate what part of the remaining problems are due to generalization issues. Including the validation towns during training increases the driving score by 20 points, improving both vehicle collisions and stop sign infractions. TF++ is close to expert level performance in this setting, ``underfitting" by 4 DS. The expert still has slightly lower vehicle collisions (Veh).

We retrain Perception PlanT \cite{Renz2022CORL} on its released dataset. It achieves a DS of 37. Perception Plant uses a handcrafted intermediate representation (bounding boxes) as visual abstraction. This makes the method more interpretable, but the downside of human designed representations are that they might miss important things. In this case, the bounding boxes do not include stop signs, leading to the method ignoring all stop signs (Stop) and incurring a large penalty for that infraction (1.87). Stop signs were not considered in the benchmark this method was developed on (Longest6).

The reproduced LAV v2 \cite{Chen2022CVPRa} also achieves a surprisingly low driving score of 27. The reason for this are its high stop sign and red light infractions. The red light infractions occur almost exclusively in Town 02 which has European style traffic lights. We evaluate on the LAV routes with additional scenarios (7,8,9,10) compared to the original benchmark. Due to the added scenarios, many traffic lights will turn yellow (and then red) just as the agent approaches an intersection. LAV v2 ignores those situations, incurring many red light infractions, bringing the overall score down. Besides the failure to adhere to traffic rules, LAV v2 achieves SotA results in route completion and vehicle collision avoidance.

\subsection{Additional Examples}
\figref{fig:TP_Recovery_OOD_2} shows another example of the importance of the target point for recovery. This time we forcefully steer the ego car onto the sidewalk, which is also an out-of-distribution situation. The target point conditioned methods (here TransFuser and TCP) extrapolate their waypoints towards the nearby target point and drive back to the center of the lane. The discrete conditioned TransFuser that does not have access to the geometric information of the target point gets stuck on the sidewalk instead.

\figref{fig:target_point_2} shows another example of harmful target point extrapolation. TCP and TransFuser both predict waypoints leaving the road in a right turn where the target point is far behind the turn. Changing the global average pooling + MLP approach in TransFuser to a transformer decoder mitigates the problem.

\subsection{Additional Experiments} 
\boldparagraph{Multimodal waypoints}
Instead of disentangling the velocity from the waypoints one could also allow the network to predict multiple sets of waypoints as is sometimes done in trajectory forecasting. To test this approach we train a model with two waypoint GRUs and a selection head to classify the better mode. During training we compute the loss as the minimum L1-loss from both predicted waypoints. The classification head is supervised with a binary cross entropy loss to classify which of the two waypoint losses has the lower L1 loss. 
\begin{table}[h]
\small
\centering
    \begin{tabular}{l | c c | c c}
        \toprule
        \textbf{Output} & \textbf{DS} $\uparrow$ & \textbf{RC} $\uparrow$ & \textbf{Veh} $\downarrow$ & \textbf{Stat} $\downarrow$ \\
        \midrule
        Multimodal WP & {47} \pmsd {8} & \textbf{96} \pmsd {1} & {1.13} & {0.20} \\
        Unimodal WP & \textbf{49} \pmsd {8} & {90} \pmsd {4} & \textbf{0.70} & \textbf{0.10} \\ 
        \bottomrule
    \end{tabular}
    \caption{\textbf{Multimodal waypoint representation.}}
    \label{tab:multimodalwaypoint}
    \vspace{-0.2cm}
\end{table}
\tabref{tab:multimodalwaypoint} shows that the representation performed 2 DS worse than using standard unimodal waypoints. This is not a big difference but the multimodal waypoints are more complex so there is not really a reason to use them.

\boldparagraph{NC conditioned AIM} To test whether recovery from the target point bias depends on the architecture, we reproduce the AIM \cite{Prakash2021CVPR} approach. Compared to TransFuser it uses no LiDAR, no auxiliary losses and no transformers, but has the same target point conditioned waypoint GRU as decoder. Like with TransFuser we run two variants, one with TP conditioning and one with NC.
\begin{table}[h]
\small
\centering
    \begin{tabular}{l | c c | c }
        \toprule
        \textbf{Cond.} & \textbf{DS} $\uparrow$ & \textbf{RC} $\uparrow$ & \textbf{Dev} $\downarrow$ \\
        \midrule
        NC & \textbf{27} \pmsd {2} & 50 \pmsd {3} & {0.96} \\
        TP & {26} \pmsd {2} & \textbf{86} \pmsd {8} & \textbf{0.00}\\
        \bottomrule
    \end{tabular}
    \caption{\textbf{Conditioning effect on AIM \cite{Prakash2021CVPR}}}
    \label{tab:aim_command}
    \vspace{-0.4cm}
\end{table}

The result presented in \tabref{tab:aim_command} also show a strong impact of TP conditioning on RC and route deviations (Dev) for the AIM architecture, suggesting that the target point bias does not depend on the particular architecture used. Consistent with \cite{Prakash2021CVPR} we also observe that AIM performs worse than TransFuser overall (-13 DS).

\boldparagraph{Target Point statistics} To show the importance of the TP for the network predictions we train a model where we input the TP as a TF decoder token so that we can analyse the attention weights. We average the attention across all layers, heads and waypoint queries. We run the model on the validation routes and observe an avg. of 25\% attention on the TP token which is 16.5x higher than uniform.
We also test how steering correlates with the target point. The sign of the steering angle (when larger than 1°), of our final model, is the same as that of the target point (TP) 92\% of the time (93\% for the expert) on the validation routes.

\subsection{CARLA Leaderboard}
The CARLA leaderboard \cite{Leaderboard2020} (we are considering version 1 in this discussion) is a test server where groups can submit agents enclosed within a docker container. These agents are then evaluated on a set of 10 routes across 2 secret towns. Each route is traversed 2$\times$ with different weather conditions. Additionally, these 20 routes are then repeated 5 times with different random seeds and average metrics across all routes are reported to the user. The CARLA leaderboard serves as a standardized evaluation platform, comparable to a test set, and some works solely rely on it to compare to other work \cite{Chen2022CVPRa, Shao2022CORL, Wu2022NeurIPS}.
Significant progress has been achieved on this benchmark. Driving scores have increased from approximately 10 DS to 70-80 DS over a span of three years.
The top performing methods on the leaderboard have released code and models alongside the release of their research papers.
In this study, we try to reproduce the top 4 reported results by submitting the released code to the leaderboard, either with the released model ($^*$) file or a retrained one ($\dagger$).

\begin{table}[ht]
\small
	\centering
    \setlength\tabcolsep{3.5pt}
	\begin{tabular}{l | c c c | c c c }
        \toprule
	    \multirow{2}{*}{\textbf{Method}} & \multicolumn{3}{c|}{\textbf{Reproduced}} & \multicolumn{3}{c}{\textbf{Reported}}\\
	    & {DS} $\uparrow$ & {RC} $\uparrow$ & {IS} $\uparrow$ & {DS} $\uparrow$ & {RC} $\uparrow$ & {IS} $\uparrow$ \\
	    \midrule
        LAV$^*$ \cite{Chen2022CVPRa} & \sout{25} & \sout{46} & \sout{0.74} & 62 & \textbf{94} & 0.64 \\
        Interfuser$^*$ \cite{Shao2022CORL} & 34 & 75 & 0.45 & \textbf{76} & 88 & 0.84 \\
        TCP $\dagger$ \cite{Wu2022NeurIPS} & 48 & 66 & 0.77 & 70 & 83 & \textbf{0.85} \\
        TF++ (ours) & {53} & 71 & 0.76 & - & - & -\\
        TransFuser $\dagger$ \cite{Chitta2022PAMI} & 55 & \textbf{90} & 0.63 & 61 & 87 & 0.71 \\
        TF++ WP (ours) & {62} & 78 & {0.81} & - & - & - \\
        TF++ WP Ens. (ours) & \textbf{66} & 79 & \textbf{0.84} & - & - & - \\
		\bottomrule
	\end{tabular}
	\caption{\textbf{CARLA leaderboard reproduced results.}} 
	\label{tab:leaderboard}
	\vspace{0.0cm}
\end{table}

\tabref{tab:leaderboard} shows significant disparities between reported results and the outcomes obtained when utilizing the official codebases. Notably, the released model of the SotA method Interfuser achieves less than half of the reported score. With the exception of TransFuser, these differences are larger than what we would expect from training or evaluation variance. The CARLA leaderboard ensures repeatability (within the bounds of evaluation variance) because one can rerun the submitted docker container. It does however not guarantee reproducibility, since released code and models can differ from what was used to achieve the reported score. Currently, it is publicly not known how to achieve scores above 60 DS, even though some groups have achieved such scores in the past. Given that the official codebases incorporate the main concepts discussed in the respective papers, these results indicate that factors not explicitly emphasized in the papers significantly influence the outcomes observed on the CARLA leaderboard.

Note, that it has been documented in \cite{Chitta2022PAMI} (Table 5) that some changes can have a significant impact on performance on the leaderboard  (+19 DS), even though the same changes do not generalize to the public towns (-2 DS on Longest6). This discrepancy can be attributed, in part, to substantial fluctuations in RC, which we don't observe in the publicly available towns. The evaluation is secret, so the exact reasons for these fluctuations are unclear. The LAV results are crossed out because the low score can be attributed to CUDA out of memory errors stemming from the released software, that appear after route 53. The CARLA leaderboard fails silently and gives 0 DS on routes where the software crashes. We contacted the organizers in this case because some of the auxiliary metrics looked unusual.

Furthermore, the auxiliary metrics on the CARLA leaderboard are typically not reported because they are known to be incorrectly computed\footnote{\url{https://github.com/carla-simulator/leaderboard/issues/117}}.
The results presented in \tabref{tab:leaderboard} were obtained with released code and models around April 2023. It should be noted that these results are subject to change if the authors decide to update their repositories.
We submit TF++ and TF++ WP (see \secref{sec:longest6_ablations}) to the CARLA leaderboard. Surprisingly, we observe a large difference of 9 DS between the representations. The reason for that difference is unclear because the effect is not reproducible on the publicly available towns.
Similar to prior work \cite{Chitta2022PAMI}, we submit an ensemble with 3 training seeds of our best model to the leaderboard. The ensemble increases the driving score by 4 points and is the best publicly available model, at the time of writing.

\boldparagraph{Discussion} The CARLA leaderboard aimed at reproducing success in past benchmarks like ImageNet for autonomous driving. It has led to similar fast progress, although the resulting methods are not always reproducible and not very well understood. There are structural changes the community could make that we think should be considered when designing future benchmarks. 

The CARLA leaderboard does not have an official validation benchmark. Since validation is fundamental to machine learning development, authors have proposed various validation benchmarks \cite{Prakash2021CVPR, Chitta2021ICCV, Zhang2021ICCV, Chen2022CVPRa, Chitta2022PAMI} using the publicly available towns.
They differ along various axes: the towns, route length, scenarios and wheathers used for validation. Based on our experience, none of the available validation benchmarks can reliably predict performance on the CARLA leaderboard. It is hard for authors to make validation routes that are well aligned because the test routes are secret. As a consequence, results on the CARLA leaderboard are often unexpected, requiring authors to run additional investigations \cite{Chitta2022PAMI}. Other authors choose not to use the leaderboard in their publications \cite{Zhang2021ICCV, Hu2022NeurIPS, Renz2022CORL, Jia2023CVPR, Zhang2023CVPR}. Submissions on the CARLA leaderboard may encounter a pending status if all servers are in use. During this project, we have encountered pending times of up to 3 weeks. Submissions themselves then take more than four days to evaluate, leading to total evaluation times of up to 4 weeks. For comparison, we can run a similar amount of simulation on our cluster in 3–6 hours, by evaluating all routes in parallel. Achieving this efficiency requires a scalable infrastructure but uses the same amount of computational resources as when evaluating routes sequentially. We encourage organizers of future testing benchmarks to release an aligned public validation benchmark that authors can use to do rapid experiments on. 

The CARLA leaderboard is a benchmark that is not associated with any dataset. Consequently, authors are required to create their own datasets for training their methods. In the early stages, the quality of these datasets, including sensor configurations, label accuracy, and scale, was often subpar. Substantial progress has been made by enhancing the quality of the data itself ~\cite{Jaeger2021}. Most papers, including this one, introduce a new dataset alongside their method. While this practice allows for innovation in dataset curation techniques, it also complicates fair comparisons between different methods. Additionally, it can obscure the source of performance improvements if the dataset is not studied as well. Although there have been some efforts in dataset curation \cite{Chen2021ICCVa, Zhang2021ICCV}, such efforts could be more fruitful if there was a standardized dataset to compare to.

\end{appendices}
\end{document}